\documentclass[lettersize,journal]{IEEEtran}
\usepackage{graphicx}
\usepackage{amsmath,amsfonts}
\usepackage{algorithmic}
\usepackage{algorithm}
\usepackage{array}
\usepackage{textcomp}
\usepackage{stfloats}
\usepackage{url}
\usepackage{verbatim}
\usepackage{graphicx}
\usepackage{cite}
\usepackage{bm}
\usepackage[usenames]{color}
\usepackage{amsthm}
\usepackage{multirow} 
\usepackage{enumitem}

\usepackage{mylatexstyle}

\usepackage{hyperref}

\begin{document}

\title{STGAN: Spatial-temporal Graph Autoregression Network for Pavement Distress Deterioration Prediction}

\author{Shilin Tong$^1$\thanks{Shilin Tong is with the Department of Computer Science, The University of Hong Kong. Email: thugh@connect.hku.hk.},~Difei Wu$^{2,*}$\thanks{Difei Wu (\textit{Corresponding Author}) and Yuchuan Du are with Key Laboratory of Road and Traffic Engineering of the Ministry of Education, Tongji University. Email: 1994wudifei@tongji.edu.cn and ycdu@tongji.edu.cn.},~Xiaona Liu$^3$\thanks{Xiaona Liu is with the Institute of IoT, Shenzhen Polytechnic University. Email: liuxiaona@szpu.edu.cn.},~Le Zheng$^4$\thanks{Le Zheng is with the School of Information and Electronics, Beijing Institute of Technology. Email: le.zheng.cn@gmail.com.},~Yuchuan Du$^2$,~Difan Zou$^{1,*}$\thanks{Difan Zou (\textit{Corresponding Author}) is with the Department of Computer Science and the Insitute of Data Science, The University of Hong Kong. Email: dzou@hku.hk}}

\markboth{IEEE TRANSACTIONS ON INTELLIGENT TRANSPORTATION SYSTEMS}%
{Shell \MakeLowercase{\textit{Tong et al.}}: STGAN: Spatial-temporal Graph Autoregression Network for Pavement Distress Deterioration Prediction}

\maketitle

\begin{abstract}
Pavement distress, manifested as cracks, potholes, and rutting, significantly compromises road integrity and poses risks to drivers. Accurate prediction of pavement distress deterioration is essential for effective road management, cost reduction in maintenance, and improvement of traffic safety. However, real-world data on pavement distress is usually collected irregularly, resulting in uneven, asynchronous, and sparse spatial-temporal datasets. This hinders the application of existing spatial-temporal models, such as DCRNN \cite{li2018diffusion}, since they are only applicable to regularly and synchronously collected data. To overcome these challenges, we propose the Spatial-Temporal Graph Autoregression Network (STGAN), a novel graph neural network (GNN) model designed for accurately predicting irregular pavement distress deterioration using complex spatial-temporal data. Specifically, STGAN integrates the temporal domain into the spatial domain, creating a larger graph where nodes are represented by spatial-temporal tuples and edges are formed based on a similarity-based connection mechanism. Furthermore, based on the constructed spatiotemporal graph, we formulate pavement distress deterioration prediction as a graph autoregression task, i.e., the graph size increases incrementally and the prediction is performed sequentially. This is accomplished by a novel spatial-temporal attention mechanism deployed by the proposed STGAN model. Utilizing the ConTrack dataset \cite{li2022contrack}, which contains pavement distress records collected from different locations in Shanghai, we demonstrate the superior performance of STGAN in capturing spatial-temporal correlations and addressing the aforementioned challenges. Experimental results further show that STGAN outperforms baseline models, and ablation studies confirm the effectiveness of its novel modules. Our findings contribute to promoting proactive road maintenance decision-making and ultimately enhancing road safety and resilience.
\end{abstract}


\section{Introduction}

Road infrastructure is fundamental to facilitating safe and efficient transportation systems, yet it faces a pervasive threat in the form of pavement distress. Characterized by cracks, potholes, and rutting, pavement distress undermines the structural integrity of roads and poses significant hazards to road users. \textcolor{black}{Under the continuous influence of the external environment and traffic loads, the pavement distress will deteriorate, escalating maintenance costs and disrupting economic and social activities. Therefore, accurately predicting the deterioration of pavement distress and conducting maintenance on time play a crucial role in improving maintenance efficiency throughout the road's life cycle.}

Accurately \textcolor{black}{predicting} pavement distress deterioration has long been a focal point in road management and maintenance research. Traditional \textcolor{black}{predicting} methods primarily utilized deterministic models, which require explicit parameterization and a defined functional form. These models are often based on mechanical theory derivation \cite{chua1994mechanistic}, empirical analysis \cite{rauhut1983damage}, or a combination of both \cite{ayed2016development}, with mechanics-empirical based pavement performance prediction models being the most established and widely used in road structure design \cite{aguib2021flexible} and maintenance decisions \cite{GARCIASEGURA2023120851}. However, pavement performance degradation is influenced by numerous factors, including meteorological conditions, pavement structure, and traffic loads, making deterministic models rely heavily on extensive observational and experimental data for parameter calibration, which limits their generalizability. On the other hand, probabilistic models represent the variability and evolution of pavement performance through statistical methods. Early approaches like gray theory \cite{wang2011pavement} and Markov-based models \cite{abaza2016back} fall under this category. Although more flexible than deterministic models, probabilistic \textcolor{black}{prediction} methods typically assume that future pavement performance changes are only related to current conditions. Despite their higher flexibility, these methods require high precision in pavement \textcolor{black}{performance} data and are thus more suited for section-based and comprehensive performance evaluation indices such as the Pavement Condition Index (PCI) and the International Roughness Index (IRI) \cite{ZHANG2024108637,KALOOP2023106007}. \textcolor{black}{However, whether it is deterministic models or probabilistic models, their prediction performance is still limited by data quality and coverage. Current prediction approaches mostly rely on annual data (e.g., LTPP dataset \cite{song2022efficient}) instead of high-frequency inspection data, making it difficult to achieve distress-level prediction and hardly considering the effect of short-term environmental changes.}

Recent advancements in pavement inspection technology have significantly enhanced the coverage and granularity of pavement distress data \cite{li2021cross}, leading to the emergence of data-driven \textcolor{black}{predicting} methods. \textcolor{black}{Additionally, the improvement of data coverage and granularity enables us to uncover the spatiotemporal evolution trends of pavement, facilitating more precise and detailed analyses, crucial for devising more effective maintenance strategies and improving pavement performance management.} Machine learning methods, widely applied in predicting pavement roughness \cite{damirchilo2021machine}, cracks \cite{ker2008development},  pavement rutting \cite{zhang2023rutting}, and even pavement temperature\cite{HATAMZAD2022108682}, as well as overall performance indicators \cite{LIN2024123696}, rely heavily on the comprehensiveness, reliability, and spatial-temporal granularity of the data samples. More recently, Li et al., \cite{li2022contrack} developed and released the ConTrack dataset, which features high spatial-temporal granularity and continuous tracking of pavement distresses, collected by crowdsensing vehicles on a road network in Shanghai, China. This dataset includes records of pavement distresses, geographic coordinates, timestamps, and environmental variables like humidity, wind speed, and precipitation levels, with a data collection interval of 1 to 7 days. Therefore, it offers great potentials to develop short-period pavement performance \textcolor{black}{predicting} models, in contrast to the prior works that typically focus on long-period prediction (e.g. annual updates).

\textcolor{black}{Short-period pavement distress prediction is a typical spatiotemporal problem, as the evolution of pavement distress is not only affected by pavement structure \& material, traffic load, and environmental factors, but it also strongly correlates with the performance changes of neighbor road sections. Several spatiotemporal models have been proposed in recent years for network-level pavement distress prediction. Wang et al \cite{wang2022spatial} conducted analysis of road traveling and its correlation with traffic flow characteristics. Chen et al \cite{chen2024large} developed a novel spatial machine-learning model to assess the road segment-based crack severity considering geocomplexity, revealing the large-scale crack deterioration assessment. Cai et al \cite{cai2023fine} adopted the emerging graph convolution networks (GCN) for short-time pavement performance prediction and demonstrated the accuracy and robustness compared with state-of-art baseline models.} However, in practical scenarios, the crowdsourced nature of the data collection means that the vehicles contribute data sporadically and randomly, resulting in sparse, irregular, and asynchronous pavement distress data. These characteristics pose significant challenges in developing effective models that can leverage this historical data for accurate prediction. As a consequence, a set of existing spatial-temporal models such as DCRNN \cite{li2018diffusion} and its variants 
\cite{li2018diffusion,wu2019graph,guo2019attention,bai2020adaptive, WEI2024111325, KONG2023110188, KHALED2022108990} \textbf{cannot} be applied. In particular, these models typically apply graph convolution operations \cite{kipf2016semi,hamilton2017inductive} in the spatial domain aggregate features from neighboring nodes and use an RNN-like \cite{hochreiter1997long} structure to update these aggregated features over time, which strictly require the data are well organized, i.e., they should be regularly collected and temporally synchronous.  \textcolor{black}{Similarly, many models heavily rely on well-organized and large amounts of temporal data, such as transformers variant models \cite{10003170}, which struggle to handle datasets with sparse temporal data and cannot effectively  incorporate spatial data connections as supplementary inputs. A notable related work is \cite{han2022asphalt}, which develops a new transformer model for predicting asphalt pavement health and is capable of handling time series data with random time differences. This work demonstrates that the new model can reduce time dependency and improve prediction performance for irregular time series. However, it focuses solely on the temporal domain and relies on sufficiently long time series data, making it unsuitable for our setting. }  Therefore, to tackle the pavement distress deterioration prediction task and similar tasks with irregular and asynchronous,  it is demanding to develop a new and more powerful spatial-temporal machine learning model.

In this work, we address this problem by proposing a novel spatial-temporal model called the Spatio-Temporal Graph Autoregression Network (STGAN) to model and perform the challenging irregular pavement distress prediction tasks, described in Section \ref{sec:prob}. The main innovations of STGAN are twofold: (1) it integrates the temporal domain into the spatial domain, constructing a unified graph for spatial-temporal data; and (2) it formulates the distress deterioration prediction as a graph autoregression problem, dynamically predicting newly added nodes (corresponding to the location and time of the distress forecast) based on historical nodes in the graph.
Specifically, each node in the spatial-temporal graph is defined by a location-time tuple, and the connections between nodes are established through a combination of hard connections and the TOP mechanism, the details are provided in Section \ref{subsection:st_graph}. Additionally, since only the time and location features are known for the newly added node, while all features are available for the historical nodes, we develop two types of feature extractions to address this inconsistency.
To enable learnable correlations between different nodes in the spatial-temporal graph, we utilize attention mechanisms based on the features of nodes and their temporal differences, which are subsequently used to perform graph convolution. The STGAN model is designed by integrating a feature extraction module, a graph convolution module, and an output MLP, which is introduced in Section \ref{subsec:entiremodel}. After outlining the architecture of STGAN, we delve into the development of its training and inference procedures for future-time pavement distress prediction, as detailed in Section \ref{subsec:train_inference}. Section \ref{sec:exp} presents our experimental findings, showcasing the superior performance of our model compared to a range of baseline methods. Additionally, we conduct a comprehensive ablation study to assess the significance of the novel modules incorporated into STGAN. Finally, Section \ref{sec:conclusion} summarizes the contributions of our work and outlines potential avenues for future research.

\section{Problem Setup}\label{sec:prob}
In this section, we will present the detailed setup of the pavement distress prediction problem and some preliminary discussions regarding the dataset and models.

\subsection{Pavement Distress Prediction Problem}

The goal of the pavement distress prediction problem is to predict future pavement deterioration given previously collected data from $N$ locations, where these historical data points may be collected at various locations and times. In the spatial domain, we have a fixed number of locations with their relative positions remaining constant. In the temporal domain, each location has a series of data collected at different times. Then we denote $X_{i, t_i^{(k)}}$ be the data collected at the $i$-th location and time $t_i^{(k)}$, where $i \in [N]$ and $k > 0$ denotes the index of the timestamp in the time series data for location $i$. More specifically, for a particular location with index $p$, the historical time series data collected at this location is denoted as $X_{p,t_p^{(1)}}, X_{p,t_p^{(2)}}, \dots$, where $t_p^{(i)}$ denotes the timestamp of the $i$-th data point at this location. Moreover, we will use a slightly different notation in the test phase:  considering a specific location from the training dataset, such as the $p$-th location, we will directly use the exact future timestamp $t'$ rather than the index of it. In this scenario, the prediction target is denoted as $X_{p, t'}$.

Then the goal this problem can be mathematically formulated as learning a function $H(\cdot)$ that takes the historical data as the input and predicts the deterioration at the location $p$ and a future time $t'$.
\begin{equation}
\big\{X_{i, t_i^{(k)}}:i\in[N], t_i^{(k)}< t'\big\}\stackrel{H(\cdot)}{\longrightarrow} X_{p,t'} \label{eq:PDFP}
\end{equation}
Note that the time variable $t'$ should satisfy $t'\ge \max_{i,k}t_i^{(k)}$ and the location variable $p$ shoud satisfy $p\in[N]$, i.e., the location should be the one that has historical information.

% The purpose of road distress prediction is to forecast the deterioration of road faults at a certain time point based on the environmental and road condition change information observed at the monitoring point $S$ and $N$ monitoring points ($X_1$-$X_n$) near $S$ in the previous period. We define the monitoring point network as a weighted directed graph $G=(V, E, W)$, where $V$ represents a set of $N$ points($|V| = N$), $E$ represents a set of edges, and $W \in R^{N \times N}$ is the weighted adjacency matrix of the graph representing the spatiotemporal relationships between points. Each point $X \in R^{P}$, where $P$ represents the number of features including environmental information, time information, spatial information, and road distress information. Assuming that $X_{i,t}$ represents the information observed at the $i_th$ monitoring point at time $t$, the objective of road distress deterioration prediction is to learn a function $H()$ that can transform the graph signals from the recent period$(t_0, t_1, t_2, t_3, \ldots ,t_n < T)$ into predicted values $Y_{i,tf}$ representing road fault deterioration over some future time periods$(tf_1, tf_2, tf_3, \ldots , tf_n > T)$, given the graph structure $G$.

% $$
% [X_{1,t_0}, X_{1,t_1}, X_{2,t_0}, X_{2,t_3}, \ldots, X_{n,t_n}; G] \stackrel{H()}{\longrightarrow} [Y_{1,tf_1}, Y_{2,tf_2}, \ldots, Y_{n,tf_n}]
% $$
\subsection{Data Analyais}

\begin{figure*}[!tb]
\centering
\includegraphics[width=0.8\textwidth]{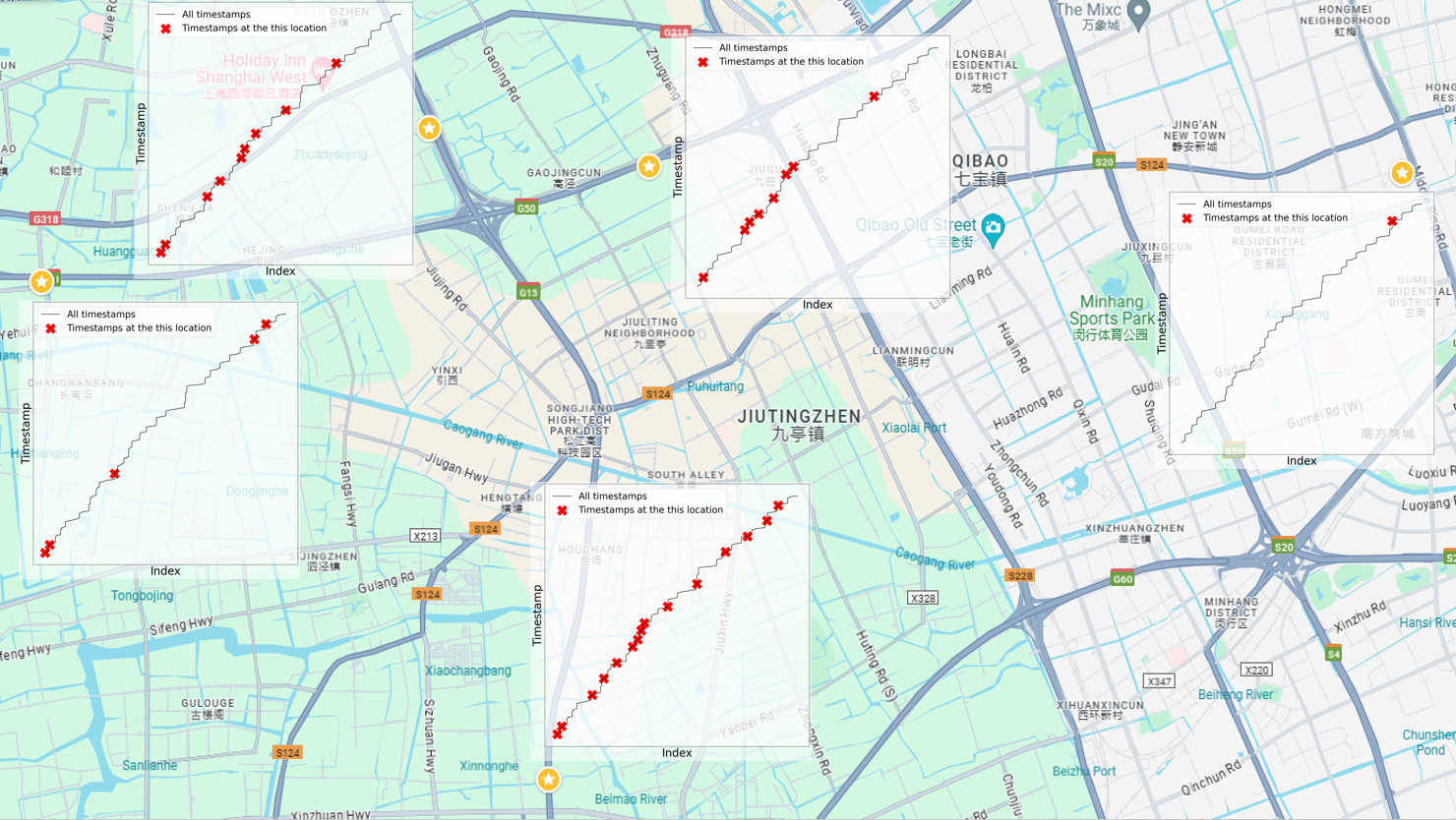}
    \centering
    \caption{Visualization of the data and timestamps at different locations in Shanghai. We selected $5$ different locations as examples. For each location, the solid line represents the timeline of all timestamps across all locations, while the red marks indicate the timestamps of the data points collected at the specific location. It can be seen that the collection patterns of timestamps vary substantially across locations, with some timestamps being quite sparse (e.g., the most right point has only one record). }
    \label{fig:issuemap}
\end{figure*}

\begin{figure}[!tb]
  \centering
  \begin{subfigure}{0.24\textwidth}
    \includegraphics[width=\textwidth]{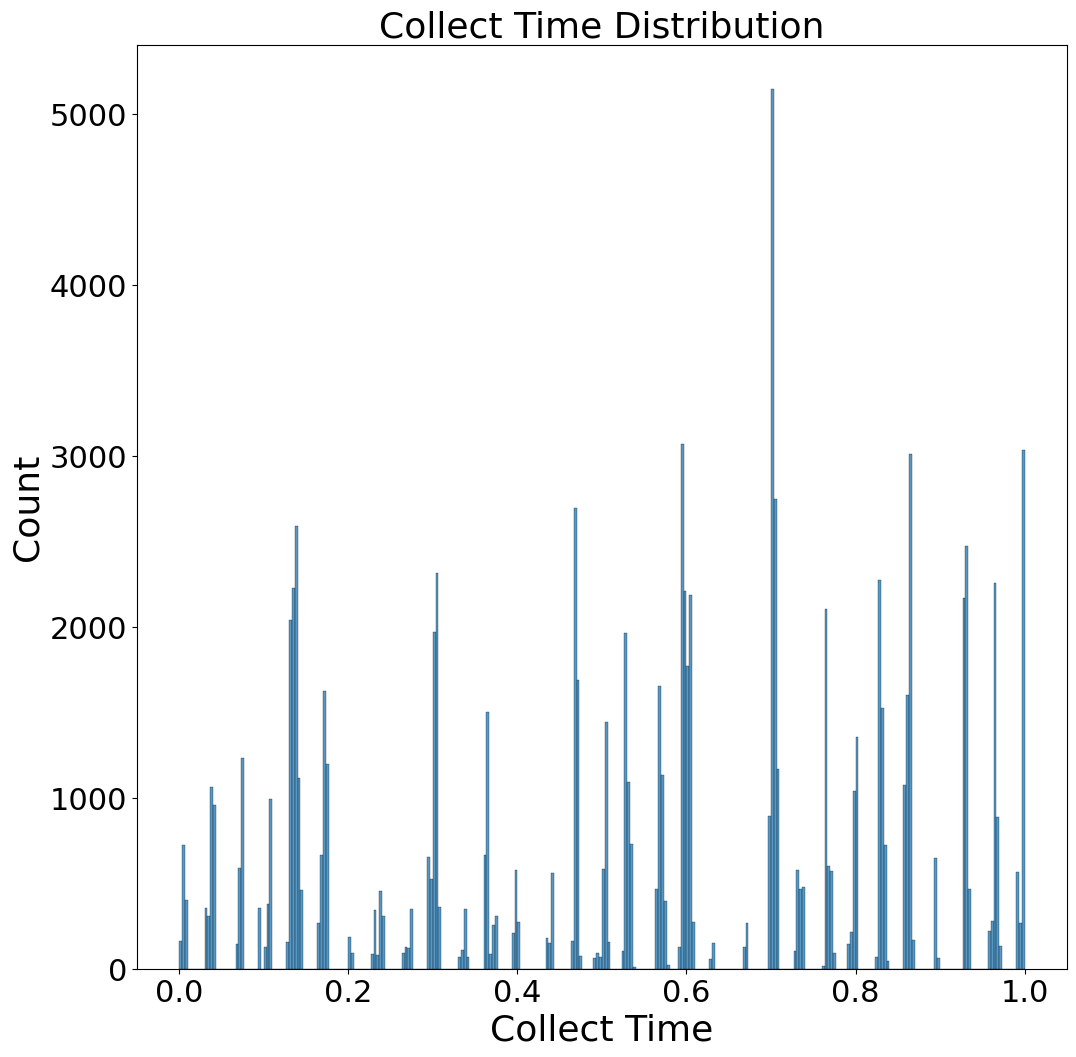}
    \centering
    \caption{Collect Time Data Distribution}
    \label{fig:issuetime}
  \end{subfigure}
  \begin{subfigure}{0.24\textwidth}
    \includegraphics[width=\textwidth]{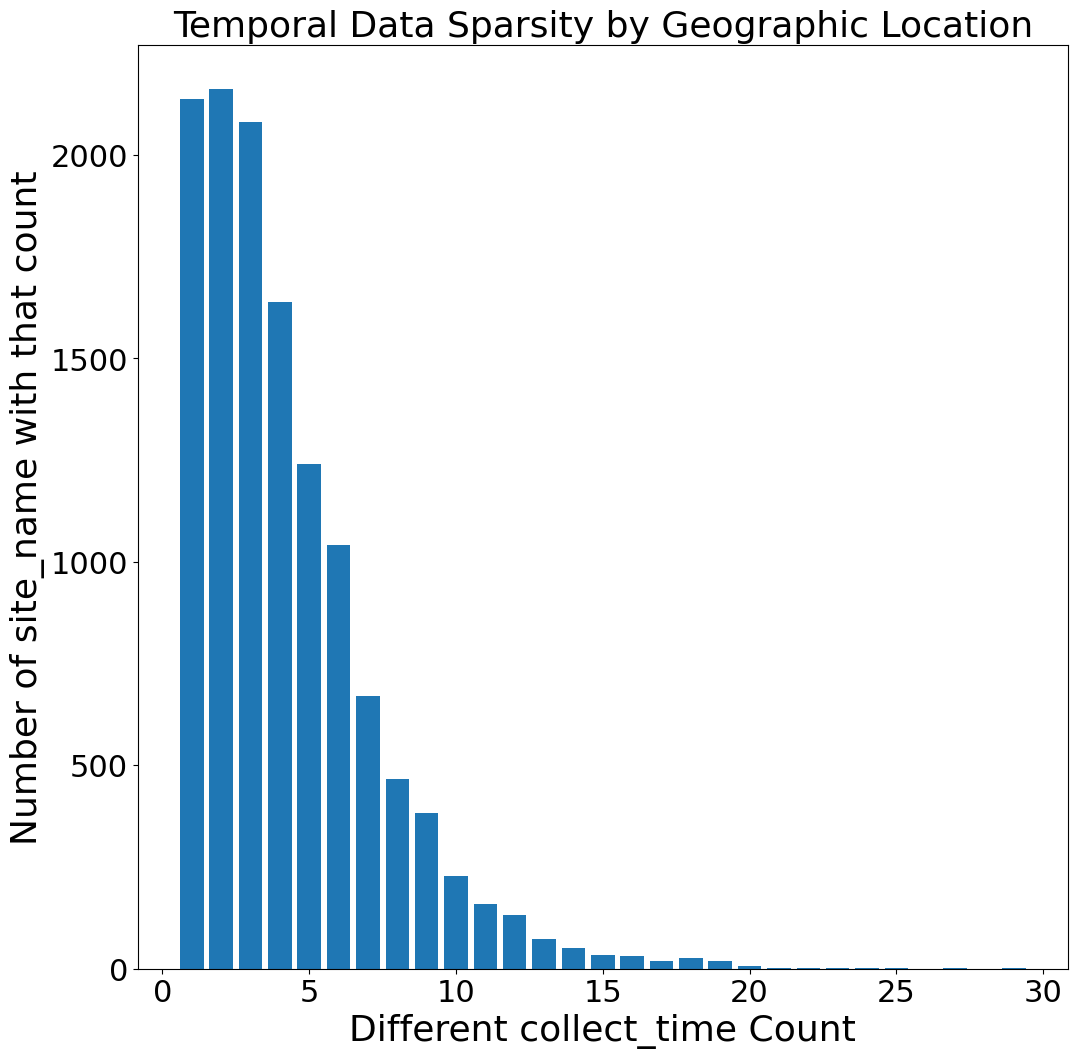}
    \centering
    \caption{Temporal Data Sparsity}
    \label{fig:issuecombine}
  \end{subfigure}
  \caption{The statistical behavior of the timestamps of all data points, which reveals the sparsity and asynchronicity properties of the pavement distress dataset in the temporal domain.}
  \label{fig:overall}
\end{figure}

As previously discussed, traffic road distress data exhibits both spatial and temporal structures. Specifically, we collect multiple data points at different times from a single location (temporal domain) and various locations (spatial domain). Following the standard data organization for spatial-temporal data, one might initially represent this information in a 2-D array, with dimensions corresponding to spatial and temporal domains. This method assumes that data collection is synchronized, meaning that at each timestamp, data features are simultaneously collected across different locations.

However, this assumption does not hold true in many practical scenarios, especially within our dataset. The road distress data is typically captured intermittently by vehicles as they travel, leading to data that is uneven, sparse, and asynchronous. For instance, the time series at different locations vary in length, frequency, and timing of data collection. In Figure \ref{fig:issuemap}, we visualize the timestamps of data points collected at $5$ locations. Then, it can be seen that the collection timestamps are substantially different for different locations, and some of them are quite sparse (i.e., even with only one timestamp). We further illustrate the statistical properties of the data points in Figure \ref{fig:issuetime}, 
including the histogram of all collection timestamps and the histogram of the number of timestamps at each location. Then it can be seen that (1) collection times are not uniformly distributed; and (2) most locations have fewer than 10 temporal data points, highlighting temporal sparsity. 

These observations underscore two major challenges in our data:
\begin{itemize}
    \item \textbf{Irregularity}: the data collection timestamps at a single location are not regular, meaning they do not follow a fixed frequency. Specifically, the timestamps $t_i^{(1)}, t_i^{(2)}, \dots, t_i^{(r)}, \dots$ do not satisfy $t_i^{(r+1)} - t_i^{(r)} = t_i^{(r'+1)} - t_i^{(r')}$ for $r \neq r'$. This is different from regular time-series data where data points are collected at consistent intervals (e.g., hourly, daily, or weekly).
    \item \textbf{Asynchronicity}: the time series at different locations have different lengths and time stamps. This implies that for different locations $p$ and $p'$, the lengths of their time series $X_{p,t_p^{(1)}}, X_{p,t_p^{(2)}}, \dots$ and $X_{p',t_{p'}^{(1)}}, X_{p',t_{p'}^{(2)}}, \dots$ are different.
    \item \textbf{Sparsity}: The time stamps are mostly sparse, the length of the time series is typically smaller than $10$.
\end{itemize}

These challenges make the data modeling substantially more difficult, and a series of previous spatial-temporal models, such as DCRNN \cite{li2018diffusion}, cannot be applied to our pavement distress forecasting problem. This exactly motivates the research objective of this paper, where we aim to build a fundamentally new model to tackle the pavement distress dataset with the aforementioned challenging patterns.

\section{Methodology}
In this section, we will introduce the main method of our study, designed to overcome the challenges associated with handling asynchronous, irregular, and temporally sparse spatial-temporal data.

\subsection{Spatial-temporal Graph For Pavement Distress Modeling}
In particular, unlike traditional spatial-temporal models that use separate modules to handle temporal and spatial data, the key motivation of our approach is to integrate these domains into a single unified graph, denoted by $\mathcal G_{ST}$. In this graph, temporal information is embedded, and each node represents a specific location and time. Consequently, predicting future events becomes a graph autoregression task. We enhance the existing spatial-temporal graph by adding a new node that corresponds to the specific location and future time we aim to predict, and then we forecast the distress value for this new node. Next, we will discuss in detail the construction of the spatial-temporal graph and the graph autoregression process.

\subsubsection{Design of the Spatial-temporal Graph}\label{subsection:st_graph}
In the spatial-temporal graph $\cG_{ST}$, each node is denoted as a location-time tuple $(i, t_i^{(j)})$, where $i$ denotes the index of the location and $t_{i}^{(j)}$ denotes the $j$-th collection time of the data in $i$-th location. In other words, unlike the graph design in other transportation problems that only treat different observation points at different spatial locations but same time as nodes in a graph, we also consider the data of the same observation point at different times as independent nodes of the same graph, thereby enhancing the comprehensiveness of our graph representation. Then, the node embedding of $(i, t_i^{(j)})$ is a collection of $d$ features that are related to the environment, time, geographical location, and pavement distress.

Once we have defined our node structure, we will move on to outline the design for the edges, aiming to capture the spatiotemporal relationships between nodes. Intuitively, nodes should be linked if they are close to each other in terms of location and if their data collection times are similar. In the following, we will introduce two connection mechanisms that will be used in this work.

\paragraph{Hard connection mechanism.}  Accordingly, we establish a directed edge pointing from a node from older time to future time, which we refer to as a hard connection between any two nodes $(i, t_i^{j})$ and $(r, t_r^{(s)})$ if the following conditions are met:
\begin{align}\label{eq:condition_connection}
|l_i - l_r|\le l_{\mathrm{res}},\text{ and }|t_i^{(j)} - t_r^{(s)}|\le t_{\mathrm{res}},
\end{align}
where $\ell_i$ and $\ell_r$ represent the locations of the two nodes, while $l_{\mathrm{res}}$ and $t_{\mathrm{res}}$ indicate the thresholds for differences in location and time stamps, respectively. 

\paragraph{TOP connection mechanism.}
However, selecting a universal threshold may result in inadequate connections between certain nodes due to the sparsity and significant differences in their time stamps. This can impede effective representation aggregation between nodes and potentially harm prediction performance. To ensure sufficient connections for all nodes, we propose the TOP compensation connection mechanism. In particular, the TOP mechanism ensures a sufficient number of connections by forcing each node to first connect to the $K$ closest neighbors, where the ranking of the neighbors is designed based on the location and time differences. Then, we then check whether there are remaining nodes that satisfy \eqref{eq:condition_connection}, which will be also connected with edges. As a consequence, the TOP connection mechanism ensures that each node will be connected to at least $K$ nodes, which leads to a dense spatial-temporal graph. A brief diagram of the proposed spatial-temporal graph is displayed in Figure \ref{fig:stgraph}.

\subsubsection{Dynamic Graph Autoregression Process}

\begin{figure}[!tb]
\centering
\includegraphics[width=0.5\textwidth]{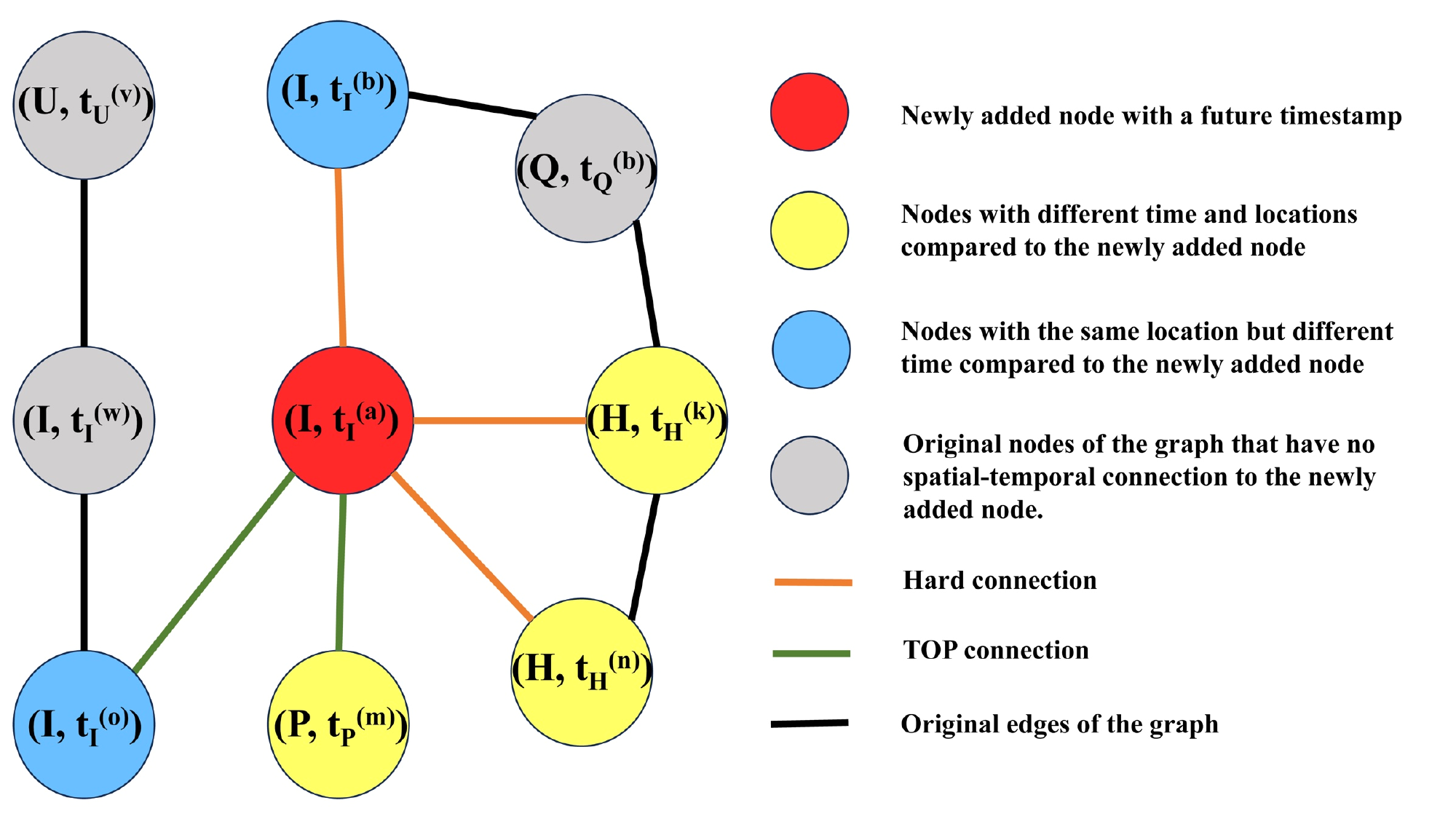}
    \centering
    \caption{Visualization of the spatial-temporal graph built based on the TOP and hard connections.  Here different nodes correspond to the different locations (e.g., $U, I, P, H, Q$) or the same location with different timestamps (e.g., $t_I^{(a)}$ and $t_I^{(o)}$). The node in red denotes the newly added one, which describes the graph expansion of the  autoregression process.}
    \label{fig:stgraph}
\end{figure}

To predict future pavement distress, a dynamic spatial-temporal graph is designed, which employs an autoregression mechanism for predictions (see Figure \ref{fig:stgraph}). 
Specifically, a period of historical data forms a graph. When predicting pavement distress at a future time $t'$ and location $p$, a new node $(p, t')$ is added to the graph. The edges connecting this new node are formed using the TOP connection mechanism. Representation aggregation is then performed on this graph, which will be discussed in detail later, to predict the pavement distress of the new node. In summary, this forms a dynamic spatial-graph autoregression model, consisting of three critical parts: graph initialization, dynamic graph expansion during the training process, and dynamic graph expansion during the testing process.

\paragraph{Graph Initialization} For the dynamic graph autoregression process, an initialized graph with a sufficient number of nodes and edges is required. Specifically, during initialization, a certain number of nodes from an earlier time period are used to build the graph. The nodes involved in the initialization are mutually visible to each other, and only hard connections are established. Enabling a node to observe data changes from both past and future nodes serves to provide data features over a certain time period for connected nodes that are temporally lagged.

\paragraph{Graph Dynamic Expansion During Training} The process of adding training data nodes is carried out incrementally based on the existing graph structure. Each newly added node naturally connects to all nodes already present in the graph, up to that time point, through both hard connections and TOP connections. Once all nodes from the training dataset have been added, a large graph is maintained for training, with the objective function defined as the average distress prediction error over the training data points. In this model, nodes in the directed graph can only predict road distress severity based on information from past time nodes, making this approach more practically valuable.

\paragraph{Dynamic Expansion During Inference} The expansion of graph in the inference period is similar to that in the training process. The primary difference is that nodes in the inference phase are added one-by-one, rather than adding all nodes to the graph simultaneously.

To better illustrate the paradigm of the proposed STGAN model, we summarize the entire process in Algorithm \ref{alg:diagram}, including initial graph construction, graph autoregression at the training phase and test phase.
\begin{algorithm}
	\caption{STGAN Paradigm} \label{alg:diagram}
	\begin{algorithmic}[1]
	\STATE \textbf{Input:} Historical data $\mathcal X=\big\{X_{i, t_i^{(k)}}\big\}$ \\
 	{\hrulefill  \texttt{Initial Graph Construction}}\ \hrulefill \\ 
        \STATE Select a set of early nodes in $\mathcal X$, denoted by $\mathcal G$.
 	\STATE Apply hard connection on $\mathcal G$ to build the initial graph. \\
        \STATE Set $\mathcal X^c=\mathcal X\backslash\mathbf \mathcal G $.\\
  	{\hrulefill \ \texttt{Training phase}}\ \hrulefill \\ 
  	\WHILE {$\mathcal X^c$ is not empty}
  	\STATE  Pick a node $Z$ from $\mathcal X^c$ based on the temporal order
    \STATE Update $\mathcal X^c\leftarrow \mathcal X^c\backslash\{Z\}$.
    \STATE Apply hard and TOP connection from $Z$ to $\mathcal G$.
    \STATE Update $\mathcal G\leftarrow \mathcal G\cup \{Z\}$.
    \ENDWHILE
    \STATE Perform training based on the entire graph $\mathcal G$.\\
    {\hrulefill \ \texttt{Test phase}}\ \hrulefill \\ 
    \STATE \textbf{Input:} Set of test nodes: $\mathcal X'=\{(p_1', t_1'),\dots,(p_r', t_r')\}$, graph $\mathcal G$
    \FOR{$Z$ in $\mathcal X'$}
    \STATE Apply hard and TOP connection from $Z$ to $\mathcal G$.
    \STATE Inference the distress information of $Z$ via STGAN.
    \STATE Update $\mathcal G\leftarrow \mathcal G\cup\{Z\}$.
    \ENDFOR
	\end{algorithmic}
\end{algorithm}

\subsection{Graph Autoregression Model Architecture}\label{subsec:entiremodel}
After constructing the spatial-temporal graph, the features of nodes are designed, and representation approaches are developed to perform the graph autoregression task.
In particular, the proposed approach involves three key components, which we outline as follows:
\begin{itemize}
\item \textbf{Feature extraction:} This component aims to develop effective modules that can better utilize the raw features of the data. Additionally, it will address scenarios where the types of data available for nodes differ between the training and testing phases. For example, nodes in the training phase may have environmental information, while nodes in the testing phase may not.
\item \textbf{Graph convolution:} This component aims to develop a module that performs information aggregation from different nodes after feature extraction. We will design a feature-aware attention mechanism to better leverage the relationships between different nodes.
\item \textbf{Graph auto-regression:} This part will utilize the representations generated through graph convolution and feature extraction to predict the distress information for each node. It can be viewed as a composition of all the modules within STGAN.
\end{itemize}

\subsubsection{Feature Extraction}

\begin{table*}[t]
    \centering
    \caption{Overview of the features}
    \label{tab:feature}
    % \resizebox{\linewidth}{!}{
    \begin{tabular}{c|c|c}
    \hline
    \textbf{Feature Type} & \textbf{Name} & \textbf{Description} \\
    \hline     
    \multirow{2}{*}{Spatial Features} & longitude\underline{ }gcj & Longitude of the observation points\\
        & latitude\underline{ }gcj & latitude of the observation points \\
    \hline
    \multirow{1}{*}{Temporal Features} & collect\underline{ }time & Time when data were collected \\
    \hline
    \multirow{8}{*}{Environmental Features}& min\underline{ }tem & Daily minimum temperature ($^\circ$C) \\
        & max\underline{ }tem & Daily maximum temperature ($^\circ$C) \\ 
        & humidity & Daily humidity (\%) \\
        & wind & Average daily Wind speed (m/s) \\
        & pressure & Daily mean air pressure (hPa)\\
        & visibility & Daily Air-visibility (km)\\
        & precipitation & Total daily precipitation (mm)\\
        & cloud & Average daily cloud cover (\%)\\
    \hline
    \multirow{7}{*}{Road Distress Features}& detect\underline{ }info & Pavement distress deterioration value and Target value \\
        & detect\underline{ }conf & Degree of confidence in distress detection \\
        & detect\underline{ }result\underline{ }type\underline{ }11 & Crack \\ 
        & detect\underline{ }result\underline{ }type\underline{ }13 & Net-crack \\ 
        & detect\underline{ }result\underline{ }type\underline{ }15 & Pothole \\ 
        & detect\underline{ }result\underline{ }type\underline{ }14 & Patch-crack \\ 
        & detect\underline{ }result\underline{ }type\underline{ }16 & Patch-pothole \\  
        \hline
    \end{tabular}
    % }
\end{table*}
% The features of each observation point data after data preprocessing are shown in Figure \ref{tab:feature}, which are divided into four types: spatial, temporal, environmental and road distress related features. The 'detect\underline{ }result\underline{ }type\underline{ }?' feature is derived by applying one-hot encoding to the road distress features in the original dataset. For instance, if 'detect\underline{ }result\underline{ }15' equals 1, it indicates that the distress type at this location is a crack.

% \hspace*{\fill}
\paragraph{Description of raw features}

In particular, the raw feature of each node comprises information from multiple aspects, including spatial features (i.e., longitude and latitude), temporal features (i.e., collection time), environmental features (e.g., temperature, wind, pressure), and road distress features (e.g., distress deterioration value, type of distress). These features and their descriptions are summarized in Table \ref{tab:feature}. It should be noted that traffic load was not taken into account in this study. This is because, within the dataset utilized, the data on this parameter are incomplete both temporally and spatially. To minimize the influence of the traffic load parameter on the prediction model and its performance, the distress data from urban roads and expressways were predominantly selected for analysis in this study. Generally, heavy-loaded vehicles are not allowed on these road sections, and the traffic volume remains relatively stable.

\paragraph{Data Pre-processing}
Noting that different features have distinct units and scalings, a standardization approach is first applied to ensure all features have the same scaling:

\begin{itemize}[leftmargin=*]
\item For all numerical features except the collection time, we leverage the training data points to standardize them to be zero-mean and unit-variance variables, i.e., let $x_1,\dots,x_n$ be the corresponding features for $n$ data points, we consider 
\begin{align*}
x_i \rightarrow \frac{x_i- \bar x}{\sigma}, 
\end{align*}
where $\bar x = \frac{1}{n}\sum_{i=1}^nx_i$ denotes the mean of features and $\sigma^2 = \frac{1}{n}\sum_{i=1}^n (x_i-\bar x)^2$ denotes the variance of features. 
\item For the features of collection time, we rescale them to $[0,1]$ by setting 
\begin{align*}
t \rightarrow \frac{t - t_{\min}}{t_{\max} - t_{\min}},
\end{align*}
where $t_{\min}$ and $t_{\max}$ denote the earliest and latest collection timestamps respectively. 
\item For categorical features, i.e., the type of pavement distress, we will use the standard one-hot encoding to map them to numerical variables. 
\end{itemize}

\paragraph{Two Types of Feature Vectors}

In general, one may feed all of these features into the model and then perform the prediction. However, considering the practical prediction process, we add a new node, corresponding to a location and a future time, and aim to predict the pavement distress in this node, the environment information and road distress information are generally missing. This implies that the input features of this new node can only encompass the spatial and temporal features, which are different from the features of other nodes. Therefore, to feasibly perform the distress prediction, we will consider two types of feature vectors for each node. In particular, at node $i$, let $X_i\in\RR^{d}$ and $X'_i\in\RR^{d'}$ be the collections of all features and only temporal and spatial features respectively, where $d$ denotes the number of all features and $d'$ denotes the number of spatial and temporal features.

A feature extractor (e.g., embedding layers) is designed to transform raw inputs into hidden representations, which are further leveraged for spatial-temporal graph representation aggregation. In this work, two-layer MLP models with ELU activation functions are considered as the feature extractor:
\begin{align}
Z_i &= \sigma\big(W_2 \sigma(W_1X_i)\big),\label{equation:featureall} \\
Z'_i &= \sigma\big(W'_2\sigma(W'_1X'_i)\big), \label{equation:featurest}
\end{align}
% \begin{align}\label{equation:featureall}
% Z_i &= \sigma\big(W_2 \sigma(W_1X_i)\big),
% \end{align}
% \begin{align}
% Z'_i &= \sigma\big(W'_2\sigma(W'_1X'_i)\big), \label{equation:featurest}
% \end{align}
where $\sigma(\cdot)$ denotes the ELU activation function, and $W_1\in\RR^{h\times d}$, $W_2\in\RR^{h\times h}$, $W'_1\in\RR^{h\times d'}$, $W'_2\in\RR^{h\times h}$ are learnable parameters that are shared for all nodes. 

\subsubsection{Attention-based Graph Convolution}

Given the representations $Z_i$ and $Z_i'$ for each node, a graph neural network model is built to perform representation aggregation between nodes. The two major components in this process involve (1) calculating edge coefficients and (2) aggregating neighboring information.

Regarding the coefficients of edges, we will follow the commonly applied attention-based mechanism in graph attention networks (GATs) \cite{velivckovic2018graph} by calculating the weight using the representations of the two nodes in the edge. However, there exists an inconsistency between the historical node and the new node in the feature space: historical nodes have all features, while the new node only has the spatial and temporal information, the environment conditions are not accessible. Therefore,  let $w_{ij}$ be the coefficient between nodes $i$ and $j$, to guarantee the symmetry in calculating the weights of edges, we will only make use of the spatial and temporal representations $Z'_i$ and $Z'_j$. Moreover, to model the similarity or correlation between nodes $i$ and $j$, it is more reasonable to look at the time difference between two nodes, as closer time typically implies stronger correlation. Therefore, let $t_{ij}$ denotes the time difference between the timestamps at nodes $i$ and $j$, the initial coefficient $w_{ij}$ can be described as
\begin{align*}
w_{ij} = \mathrm{LeakyReLU}\big(W_e[Z'_i||Z'_j||[t_{ij}]]\big)
\end{align*}
where $||$ denotes the vector concatenation, and $W_e$ is a learnable weight matrix. Then in our setting, each node will be interacting with all of its neighborhoods to perform representation aggregation. Therefore, based on the calculated initial coefficient $w_{ij}$ for any pair of nodes, we will further apply the graph structure information to perform the masked attention, i.e., only calculating the coefficient between the node and its neighborhood. Mathematically, let $N_i$ be the (parent) neighborhood set of  node $i$, the attention coefficient $a_{ji}$ has the following formula:
\begin{align}\label{eq:attention_coefficient}
a_{ji} = \frac{\exp(w_{ji})}{\sum_{b\in N_i}\exp(w_{bi})}.
\end{align}
In addition, we will employ a multi-head attention mechanism, where multiple spatial-temporal correlation coefficients are calculated. This implies for each edge $e_{ij}$, we will use a number of learnable weight matrices $W_e^1,\dots,W_e^H$, where $H$ denotes the number of heads, and then get a set of attention coefficients $a_{ij}^{1}, \dots, a_{ij}^H$ accordingly. These coefficients will be leveraged in the representation aggregation step.

Regarding the representation aggregation, i.e., aggregate the information from neighborhoods to update the representation of the current node, we will seek to use the full representation to collect distress information as much as possible. Notably, this will not lead to information leakage issue as the directed graph ensures that one can only aggregate the distress information from the nodes with earlier timestamps. Therefore, the representation aggregation on the node $i$ and the attention head $h$ can be mathematically formulated as follows:
\begin{align}
\tilde Z_i^{(h)}  &= \sum_{b\in N_i, b\neq i}a^{(h)}_{bi}Z_b + a^{(h)}_{ii}\cdot Z_i', \label{equation:gcn_convolution}\\
\tilde Z_i &= \tilde Z_i^{(1)}||\tilde Z_i^{(2)}||\cdots||Z_i^{(H)}. \label{equation:nultiple_head_gcn}
\end{align}
Then, we will further apply an MLP to further map the high-dimensional representation $\tilde Z_i$ back to the same dimension of $Z_i$ to obtain the representations in the first layer, i.e.,
\begin{align}
Z_i^1 = \mathrm{ELU}(W_o \tilde Z_i).
\label{equation:Gcov_MLP}
\end{align}
where $W_o$ is the weight matrix that will be optimized during the training.

\subsubsection{Spatial-Temporal Graph Auto-regression Network}
We have introduced the feature extraction and graph convolution module. Then, the forward propagation of the entire model, denoted as spatial-temporal graph auto-regression network (STGAN), can be performed by combining (1) an input MLP that performs feature extraction, (2) multiple-layer graph convolution; (3) and an output MLP (or decoder). In particular, the graph convolution operation mentioned above is a combination of \eqref{equation:gcn_convolution}, \eqref{equation:nultiple_head_gcn} and \eqref{equation:Gcov_MLP}. For the last GConv layer, we do not include the MLP layer but will merge it into the final output MLP module. Thus the output of the last graph convolution layer is $\{\tilde Z_i^L\}_{i=1,\dots,N}$ rather than  $\{ Z_i^L\}_{i=1,\dots,N}$. For the remaining layers $\ell \ge 1$, the difference to the graph convolution operation in the first layer is that we will use one set of representations rather than two. In particular, given the hidden representations $\{Z_i^l\}_{i=1,\dots,N}$, the graph convolution step calculates
\begin{align*}
{Z}_{i}^{l+1}  &= \mathrm{ELU}\bigg(W_{o}\bigg[\sum_{b\in N_{i}} a^{l,(1)}_{bi} Z_{b}^{l} ||\cdots||\sum_{b\in N_{i}, b\neq i} a_{bi}^{l,(H)} Z_{b}^{l}\bigg]\bigg),
\end{align*}
where $a^{l,(h)}_{bi}$ is the attention coefficients calculated via \eqref{eq:attention_coefficient} based on the representations $\{Z_i^l\}_{i=1}^n$.

\begin{figure*}[!tb]
  \centering
\includegraphics[width=0.75\textwidth]{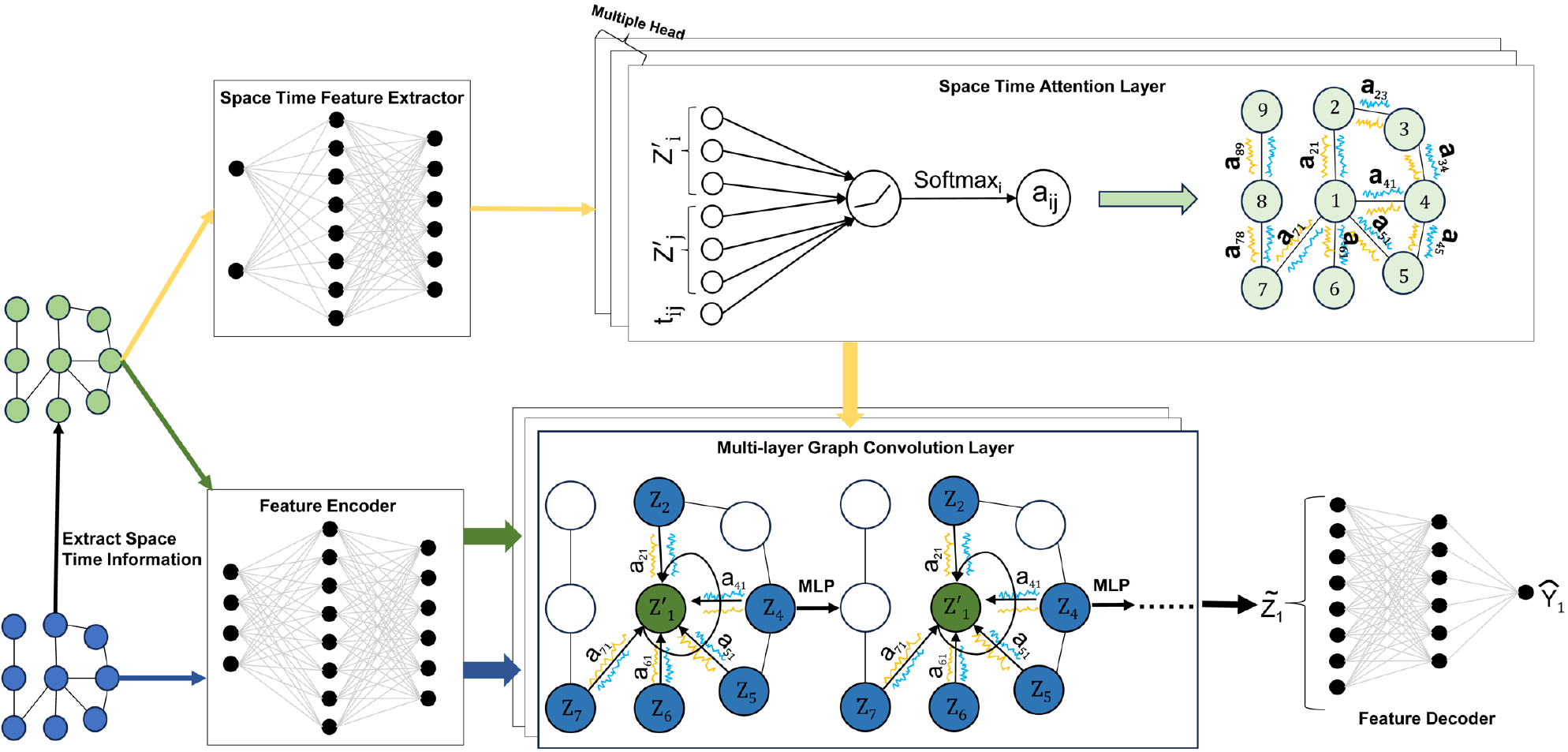}
  \caption{Diagram of the spatial-temporal graph autoregression network model.}
  \label{fig:stgan_archi}
\end{figure*}

In summary, the entire forward propagation of STGAN is displayed as follows:
\begin{align*}
&\{X_i, X_i'\}_{i=1,\dots, N}\rightarrow\text{Input MLP} \rightarrow \{Z_i, Z_i'\}_{i=1,\dots,N}\\
&\rightarrow \text{GConv}\rightarrow\{Z_i^{1}\}_{i=1,\dots,N}
\rightarrow \underbrace{\text{GConv}\rightarrow\cdots \text{GConv}}_{\text{$L-1$ layers}}\\
&\rightarrow \{\tilde Z_i^{L} \}_{i=1,\dots,N}\rightarrow  \text{Output MLP}\rightarrow \{\hat Y_i \}_{i=1,\dots,N}.
\end{align*}
Here the input and output MLPs are operated in a node-wise manner and $\hat Y_i$ denotes the predicted pave distress value at the node $i$. The detailed illustration of the entire STGAN model is shown in Figure \ref{fig:stgan_archi}.
 % The primary reason for distinguishing them in our equations is to maintain consistency with the first layer of GConv, which uses features from two different feature extractors. Therefore, $Z'$ represents the features of the target points, while $Z$ represents the features of the other points.

% \begin{align*}
% {Z'}_{i}^{l+1}  &= \mathrm{ELU}(W_{o}(\sum_{b\in N_{i}, b\neq i} a^{1}_{bi} Z_{b}^{l} + a_{ii}^{1}\cdot{Z'}_{i}^{l}\quad||\sum_{b\in N_{i}, b\neq i} a^{2}_{bi} Z_{b}^{l} + a_{ii}^{2}\cdot{Z'}_{i}^{l}\quad||\cdots||\sum_{b\in N_{i}, b\neq i} a_{bi}^{h} Z_{b}^{l} + a_{ii}^{h}\cdot{Z'}_{i}^{l}))
% \end{align*}

\subsection{Model Training and Inference}\label{subsec:train_inference}
\paragraph{Model Training} In the training period, the spatial-temporal graph $\mathcal G_{ST}$ consists of the initial nodes and all training nodes. Then, let $\{(X_i,X_i'; Y_i)\}_{i=1,n}$ be the training set, where $i$ denotes the index of the data that is determined by the location and collection time, $X_i$ denotes all input features,  $X_i'$ is a subset of $X_i$, which only covers the spatial and temporal features. Then, our spatial-temporal graph neural network model will take the graph $\mathcal G_{ST}$ and all training features as inputs. Mathematically, we denote 
\begin{align*}
F_{W}(\mathcal G_{ST}; X_1', X_1,X_2',X_2,\dots, X_n',X_n)\rightarrow\hat Y_1,\dots,\hat Y_n\tag{STGAN Function}
\end{align*}
as the function of the STGAN model, where $W$ denotes the model parameter and the output is the collection of distress predictions for all training nodes. Here we would like to emphasize that $\mathcal G_{ST}$ is a directed graph, thus the information in $X_i$ will not be leaked to the prediction of $Y_i$ as our graph convolution operation will leverage $X_i'$
to predict $Y_i$. Then, the model parameter $W$ will be trained based on the following mean absolute value (MAE) loss function:
\begin{align*}
L(W) = \frac{1}{n}\sum_{i=1}^n |\hat Y_i - Y_i| \tag{Training Loss}.
\end{align*}

\paragraph{Model Inference} 
In the inference period, the distress will be predicted one-by-one. In particular, given the current spatial-temporal graph $\mathcal G_{ST}$ with $n$ training nodes, a new node with index $n+1$, corresponding to the target location and future time, will be added, leading to a larger graph $\mathcal G_{ST}'$. Then based on the features of all existing nodes and the spatial and temporal features $X_{n+1}'$ of the new node, let $\hat W$ be the learned model parameter, the inference of the STGAN model can be formulated as follows:
\begin{align*}
g_{\hat W}(\mathcal G_{ST}'; X_1', X_1,X_2',X_2,\dots, X_n',X_n, )\rightarrow \hat Y_{n+1}\tag{STGAN Inference}.
\end{align*}
When predicting the next future timestamp, i.e., $n+2$, we can either (1) use the true distress information for the node $n+1$ if they can be collected; (2) use the predicted distress information for the node $n+1$ if no physical collection is performed; and (3) ignore the node $n+1$, but directly predict the pavement distress for the node $n+2$ by treating it as a new node to the training spatial-temporal graph $\mathcal G_{ST}$.

\section{Experiments}\label{sec:exp}

\subsection{Dataset Description}

\begin{figure}[!t]
    \centering
    \includegraphics[width=1\linewidth]{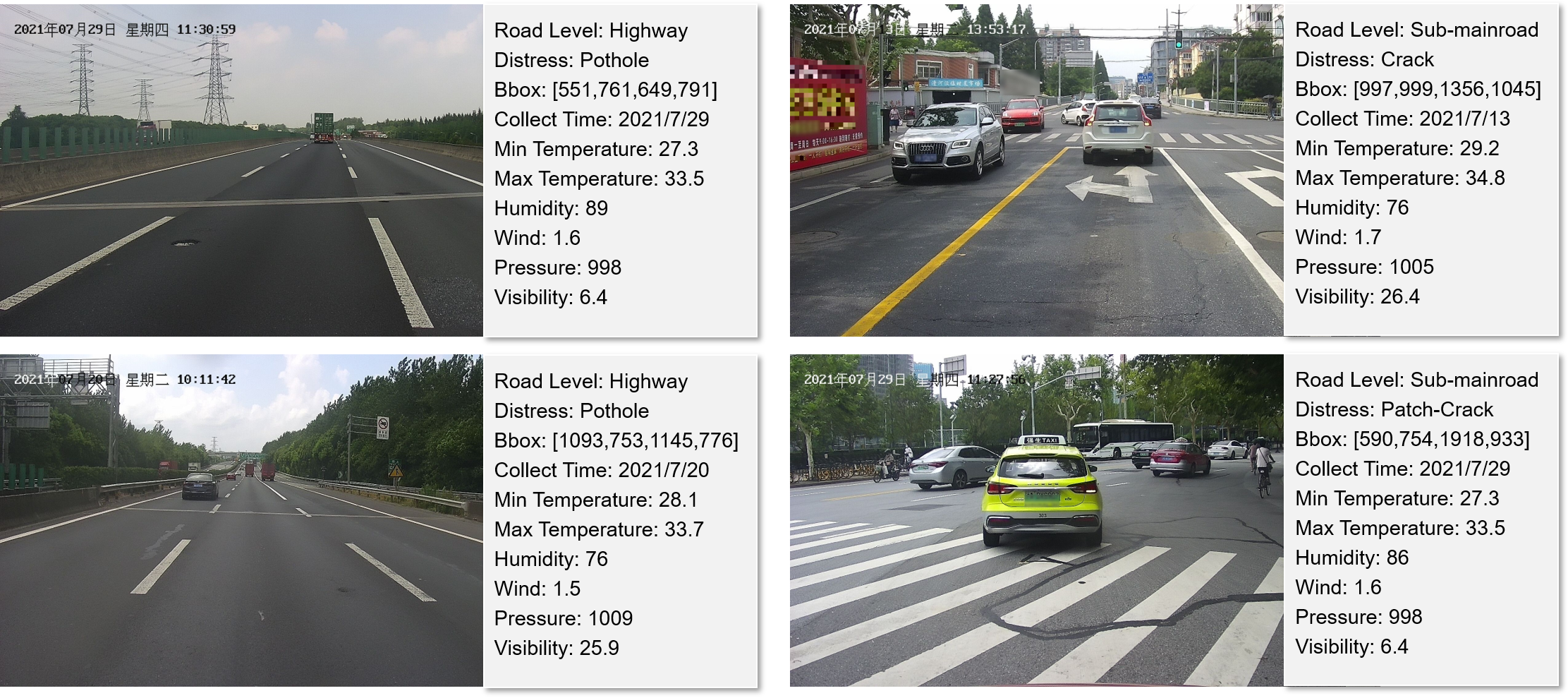}
    \caption{Samples in the ConTrack Dataset. The orginal data is represented as a image collected by the camera on a car and  environment conditions collected by the sensors. }
    \label{fig:enter-label}
\end{figure}

The ConTrack dataset utilized in this study constitutes a continuous and high-frequency observational dataset of pavement distress within regional road networks. The dataset comprises approximately 47,000 instances of pavement distress, with over 14,000 of them being continuously tracked. Pavement distress data were collected by recorders on various public vehicles, including inspection vehicles, buses, and logistic trucks. This collection encompasses original images of pavement distress, along with identified distress categories, bounding boxes, collection timestamps, and other pertinent information. Several samples are illustrated in Figure \ref{fig:enter-label}. Within this dataset, bounding box information is utilized for calculating the dimensions of each pavement distress. Specifically, distress categories and bounding boxes are identified using the EfficientDet deep learning model, achieving an overall recognition accuracy of 86\%. The identified distress categories include crack, patch-crack, pothole, patch-pothole, net-crack, and patch-net-crack, covering typical types of pavement distress. The collection time for distress image information primarily falls between 10:00 and 14:00 to ensure adequate lighting and to avoid traffic peaks in the early morning and evening, thereby reducing the impact of surrounding vehicle obstruction.

The dataset spans from April 2021 to June 2022, encompassing over 350 kilometers of urban roads and highways in Shanghai, China, with observations recorded at a day-level frequency. Furthermore, the dataset incorporates meteorological data and geographic information for each distress instance, facilitating the detailed analysis of distress evolution, temporal and spatial distribution, and its correlation with meteorological conditions. After data cleaning, standardization of continuous feature values and one-hot encoding on fault type features, we selected 11 features of four dimensions as observation point data, including time features, location features, environmental features, and road distress features. The preprocessed dataset contains a total of 17,200 nodes.

% \subsection{Experimental Setting}

% we conduct comparative experiments by juxtaposing our model STGAN with four frequently employed models in traffic prediction tasks, some of which are adapted according to Contrack dataset and problem-specific features: TOP-MLP, GCN, GCN-MLP, and GAT. Subsequently, we analyze the results of ablation experiments to discuss the roles of three crucial components or elements within the STGAN model: the TOP connection, spatiotemporal attention mechanism, and temporal delta features. Finally, we perform comparative experiments on hyperparameters such as the number of convolutional layers and attention heads within STGAN, analyzing the impact of different parameter values on the results. All these methods are based on neural networks, implemented using PyTorch, and employ the Adam optimizer with learning rate annealing.
\subsection{Baseline Methods and Evaluation Criteria}
We first present the baseline methods and evaluation criteria that will be used for comparison. Note that the standard spatial-temporal graph networks, e.g., DCRNN  cannot be applied to our dataset, we will design the baseline algorithms by using simple MLP, GCN, and their combinations on our designed spatial-temporal graph (please see Section \ref{subsection:st_graph}).

\subsubsection{Baseline Methods}
In this section, we will introduce the baseline methods used to demonstrate the performance of the developed STGAN model. Notably, since existing spatial-temporal models cannot be applied to our task, we will ablate the developed STGAN model to create the baseline methods. Specifically, we will consider the following four baseline models. These models will help us gain a comprehensive understanding of the various modules in our STGAN model, including graph construction, feature extraction, and graph convolution.
In particular, the considered baseline methods are organized from simple to complicated, including TOP-MLP, GCN, GCN-MLP, and GAT. 
\begin{itemize}[leftmargin=*]
\item The \textbf{TOP-MLP} model is the simplest one that does not use graph convolution operations. We have incorporated the spatial information into MLP by using the proposed TOP connection mechanism to encode the spatial information as the features of each data point.
\item	The \textbf{GCN} model is the standard one that applies the graph convolution operation with a fixed graph Laplacian, which is generated based on the connection mechanism proposed in Section III.A.
\item	The \textbf{GCN-MLP} model is the combination of GCN and MLP model, which leverages GCN model to extract features (instead of the raw features used in MLP) and then uses the MLP model to perform the prediction.
\item	The \textbf{GAT} model is a more advanced graph network model that applies the attention mechanism to perform the graph convolution. We use the standard attention mechanism in the original GAT paper \cite{velivckovic2018graph}, which is different from our specifically designed attention calculation mechanism in Section III.B.
\end{itemize}

\begin{figure*}[ht]
    \centering
    \begin{subfigure}{0.49\textwidth} % 设置第一张图片占总宽度的45%
        \centering
        \includegraphics[width=\textwidth]{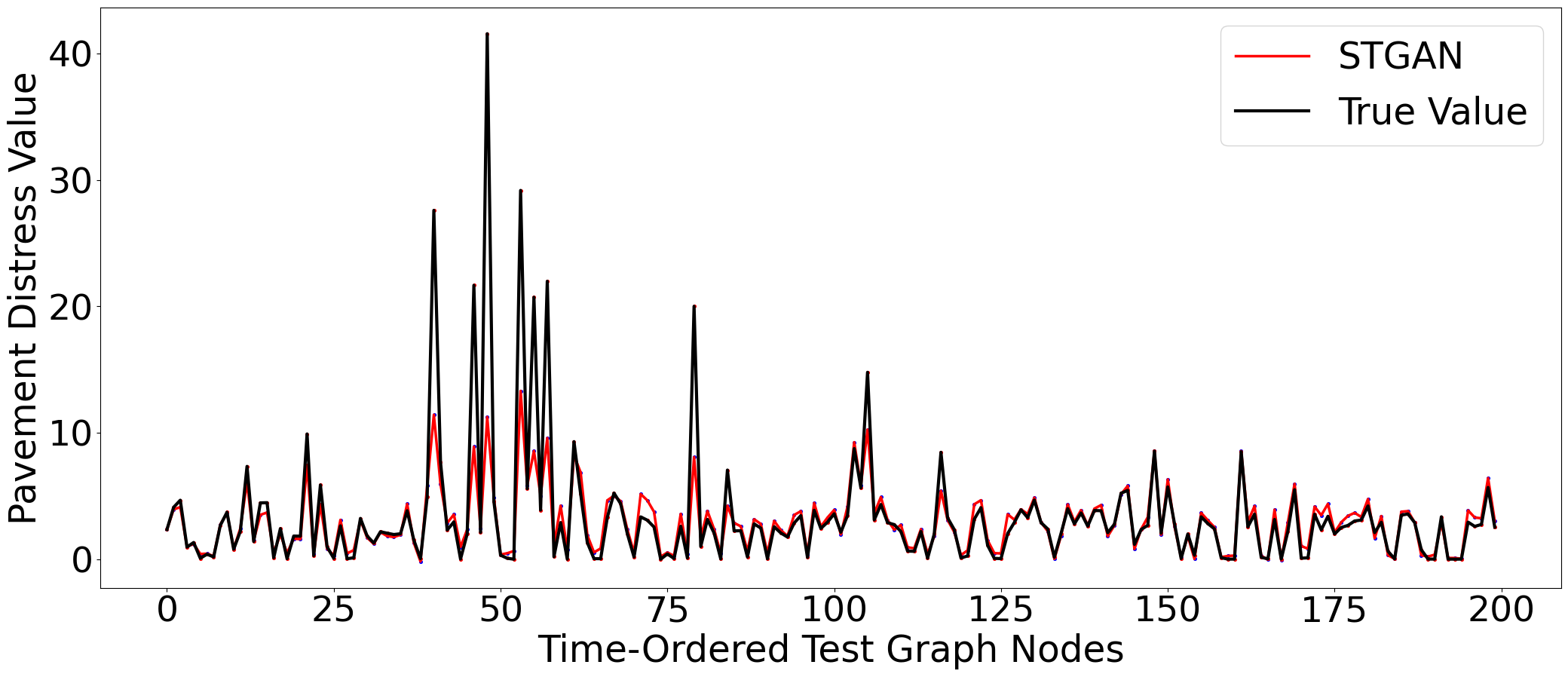}
        \caption{0-2000 Data}
        \label{subfig:STGAN_pre_true_a}
    \end{subfigure}
    \begin{subfigure}{0.49\textwidth} % 设置第二张图片占总宽度的45%
        \centering
        \includegraphics[width=\textwidth]{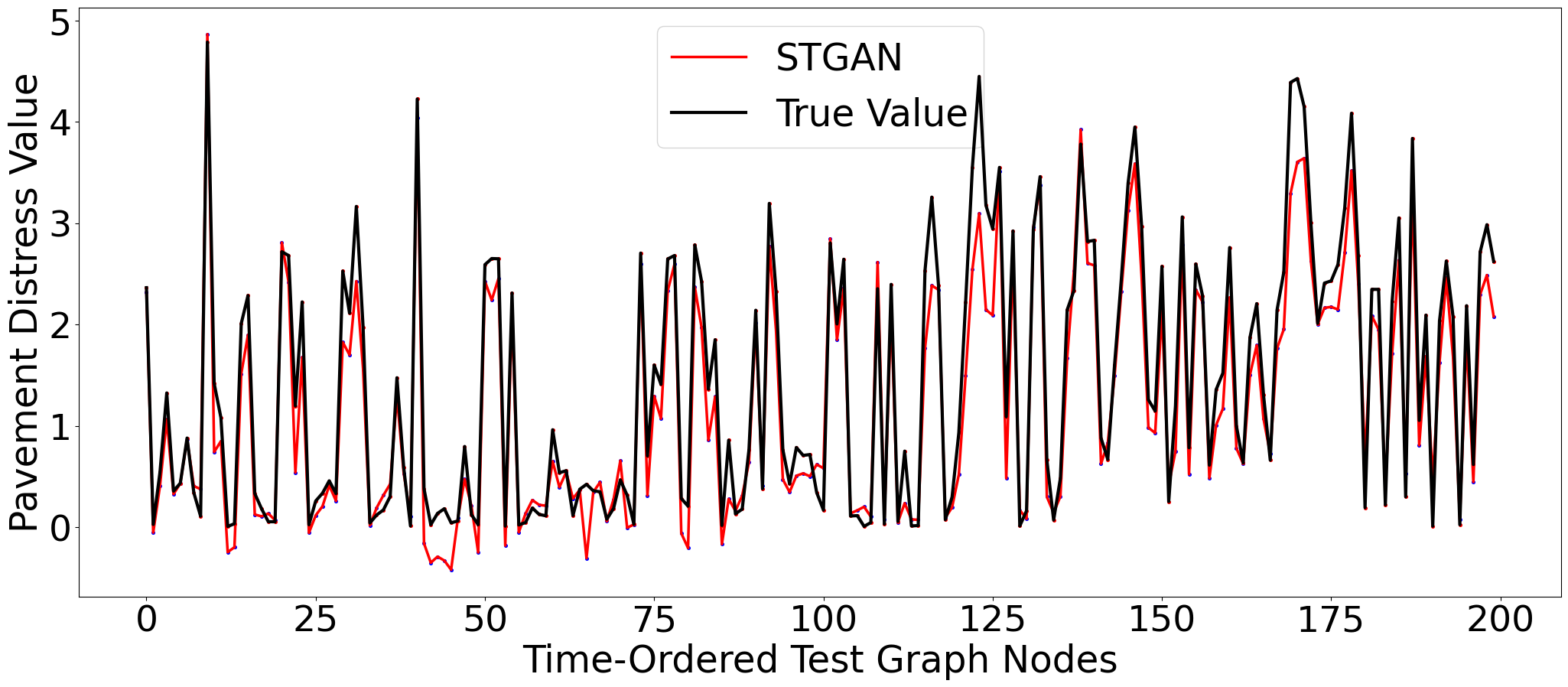}
        \caption{5000-7000 Data}
        \label{subfig:STGAN_pre_true_b}
    \end{subfigure}
    \begin{subfigure}{0.49\textwidth} % 设置第三张图片占总宽度的45%
        \centering
        \includegraphics[width=\textwidth]{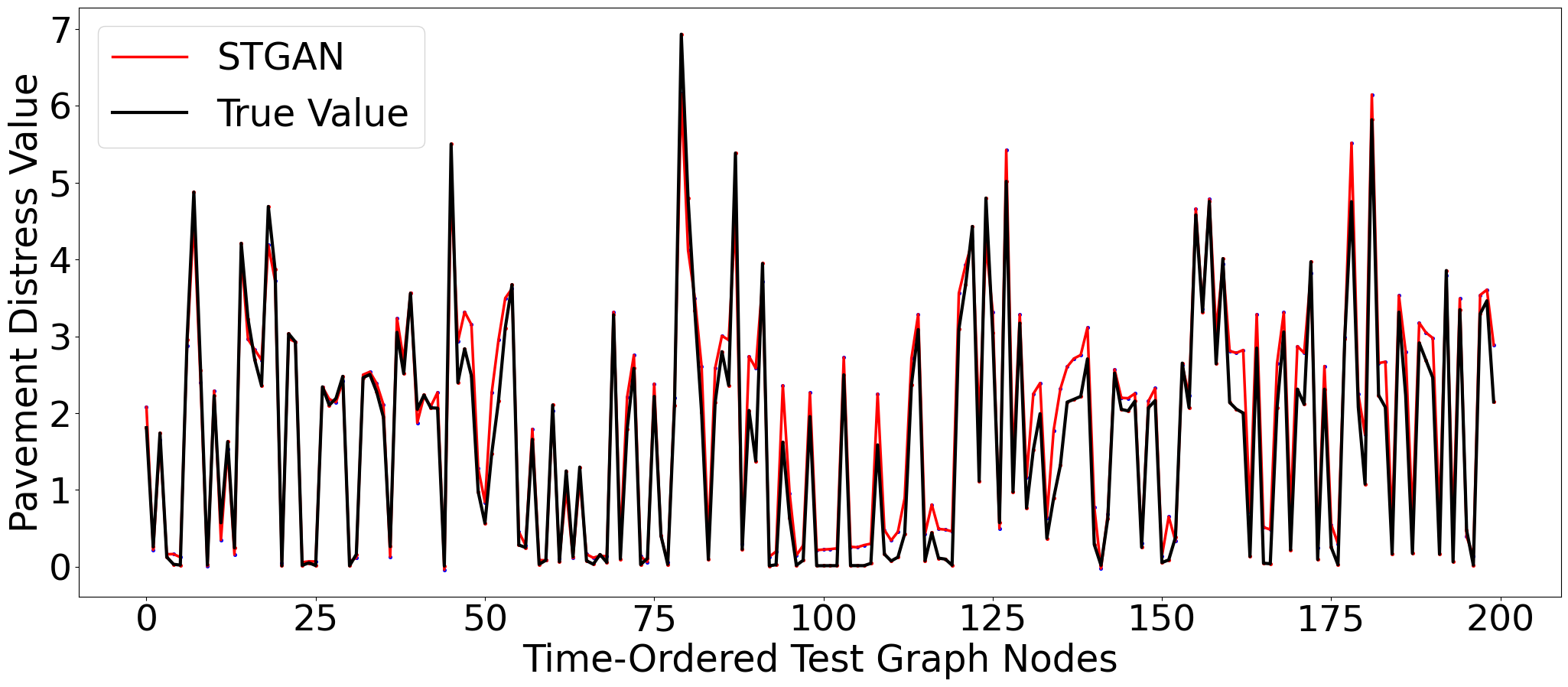}
        \caption{10000-12000 Data}
        \label{subfig:STGAN_pre_true_c}
    \end{subfigure}
    \begin{subfigure}{0.49\textwidth} % 设置第四张图片占总宽度的45%
        \centering
        \includegraphics[width=\textwidth]{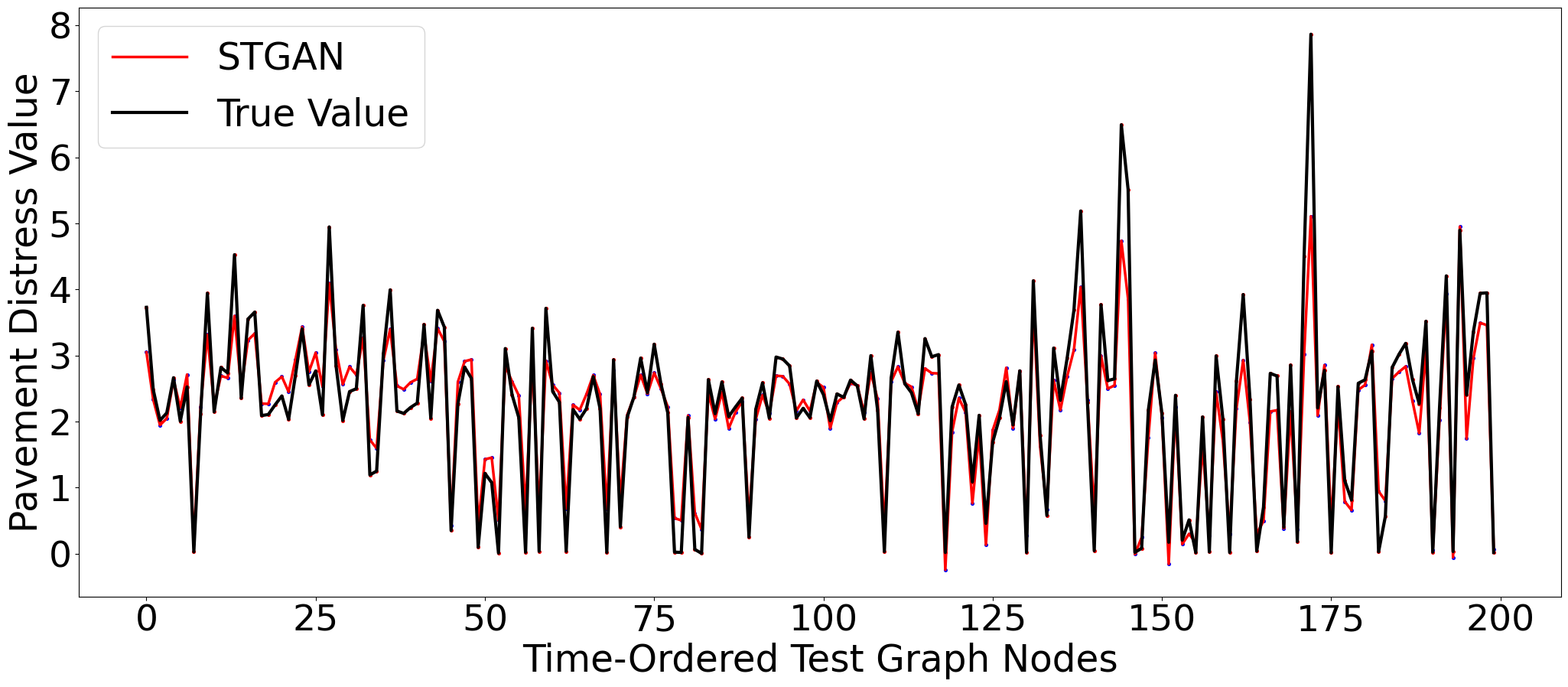}
        \caption{15200-17200 Data}
        \label{subfig:STGAN_pre_true_d}
    \end{subfigure}
    \caption{Performance of the STGAN model in different periods.}
    \label{fig:STGAN_pre_true}
\end{figure*}
\subsubsection{Evaluation Criteria}
The distribution of pavement distress values varies over different time periods. To explore the model's performance across different distributions of target values, we selected four distinct data segments from the dataset. In each data segment, 10\% (200 data points), 70\% (1400 data points) and the remaining 20\% (400 data points) of all data are used for graph initialization, training, and testing, respectively. Notably, due to the dynamic graph expansion based on time series, we do not use a validation set as it would create a significant temporal gap between the training and testing sets, potentially reducing the predictive accuracy of the testing set. Therefore, we evaluate the performance of different models solely on the training and test datasets, resulting in training error and test error. Particularly, we consider the following three evaluation metrics, including Mean Absolute Error (MAE), Mean squared Error (MSE), and Root Mean Squared Error (RMSE). Their formula are given as follows:
\begin{align}
\text{MAE} = \frac{1}{n} \sum_{i=1}^{n} |y_i - \hat{y}_i|, \notag\\
\text{MSE} = \frac{1}{n} \sum_{i=1}^{n} (y_i - \hat{y}_i)^2, \tag{\text{Evaluation Metrics}}\\
\text{RMSE} = \sqrt{\frac{1}{n} \sum_{i=1}^{n} (y_i - \hat{y}_i)^2}.\notag
\end{align}
where $\{\hat y_i\}_{i=1,\dots, n}$ and $\{y_i\}_{i=1,\dots,n}$ denote the predicted values and true values respectively.

In our experiment, we will also involve the multi-class prediction to predict the distress level. To quantify the prediction accuracy, we will use AUC, i.e., the area under the Receiver Operating Characteristic (ROC) curve, which plots the True Positive Rate (TPR) against the False Positive Rate (FPR) at various threshold settings. In particular, a larger AUC (close to 1) implies that the model can correctly predict the label with a higher probability. Instead, when AUC is approaching 0.5, this implies that the predictor behaves similarly to the random guess (i.e., the performance is bad).

\subsection{Experimental Result}

\paragraph{Experiment Setup.}
We first explain our experiment setup as follows:
\begin{itemize}[leftmargin=*]
\item Regarding the \textbf{graph construction}: the number of TOP connection is set as $5$ to balance the complexity of the graph and the sufficient connection between nodes. We have performed the experiments for different choices ($0-10$) where it has been found that smaller number leads to insufficient training and larger ones lead to slight overfitting.
\item Regarding the \textbf{STGAN model architecture}, the feature extractor and graph convolution module are designed as follows:
\begin{itemize}[leftmargin=*]
\item \textbf{Feature extractor:} We use a three-layer MLP with hidden dimensions $18$, $128$, and $256$. The first dimension is determined based on the number of features; the second and the third ones are selected by balancing the model capacity and overfitting, which is suggested by multiple trials on different choices (e.g., $64,128,256,512$).
\item \textbf{Graph convolution module:} We use multi-head attention and multiple-layer GAT. The hidden dimension is kept to be $256$ and numbers of layers and heads are also selected by multiple trails (see Section IV.G for ablation study).
\end{itemize}
\item Regarding the \textbf{model training}, we use default ADAM optimizier with learning rate $0.004$, running for $200$ epochs. This is selected by grid search around the typical configuration.
\item Regarding the \textbf{data preprocessing}, we selected four continuous data segments from the dataset, ordered in terms of their timestamp (the adjacent two data points can be from different locations),  with $2000$ data points in each segment: $0-2000$, $5000-7000$, $10000-12000$, and $15200-17200$. This is designed to investigate the  prediction performance of STGAN across various time periods and data distributions.
\end{itemize}

\paragraph{Predict the distress value.}
The comparison between the prediction and the ground-truth values is visualized in Figure \ref{fig:STGAN_pre_true}, including the predictive performance of STGAN on four different data segments. 

It can be observed that within the range of small deterioration values, the model fits well and achieves a relatively high accuracy. However, for certain substantially high deterioration values, the prediction accuracy will be downgraded. This is because extremely severe pavement distress is less common in the dataset, and the transition to such severe conditions from milder distress occurs rapidly. In most cases, these severe conditions are caused by rare events related to weather or road construction issues, making it challenging for the model to predict such variations accurately.

\begin{figure}
  \centering
  \includegraphics[width=0.5\textwidth]{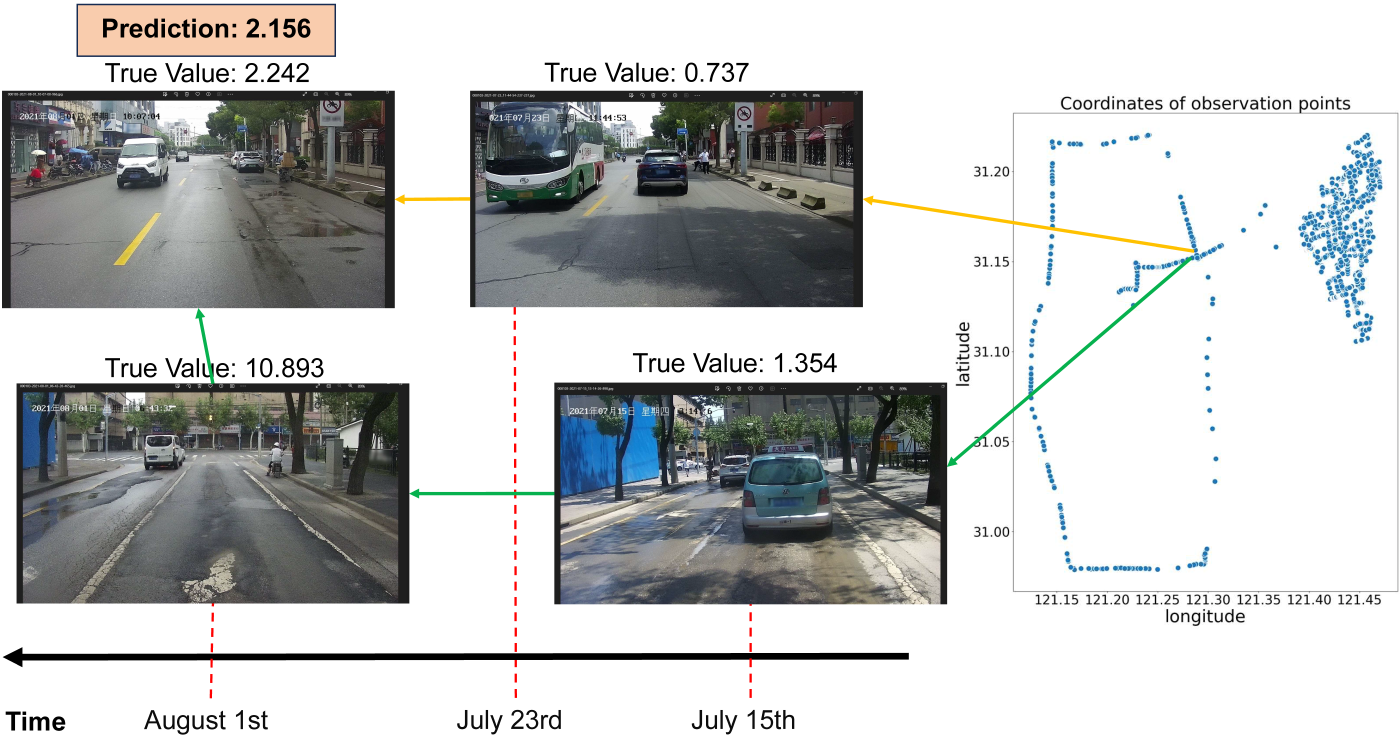}
  \caption{Illustration of the  Spatial-temporal connections between the distresses at different locations and different timestamps.}
  \label{fig:actualconnection}
\end{figure}

As illustrated in Figure \ref{fig:actualconnection}, the dataset displays two occurrences of traffic pavement distress, both of which worsened on August 1st. By capturing the real spatial-temporal relationships between pavement distresses, STGAN effectively predicts the extent of road distress deterioration at a specific future time. This prediction is based on the previous condition of the distress and the deterioration patterns of nearby distresses.

\paragraph{Predict the distress level.}
In practice, it is more common to consider the distress level rather than the exact value when evaluating road conditions. Therefore, we assess the performance of STGAN in handling the task of predicting distress levels, as described below:
\begin{enumerate}
    \item \textbf{Healthy}, where the fault value ranges from 0 to 1.
    \item \textbf{Good}, with fault values between 1 and 5.
    \item \textbf{Severe}, encompassing fault values from 5 to 10.
    \item \textbf{Very severe}, indicating fault values exceeding 10.
\end{enumerate}

The goal of categorization is to address imprecise predictions of high fault values while maximizing the model's accuracy in predicting low fault values, ultimately enhancing STGAN's practicality. When the model predicts a change in road distress classification from class 3 to class 4, an alert is sent to relevant transportation departments, prompting them to repair the road surface, mitigate traffic hazards, and prevent accidents. 

\begin{figure}[!t]
  \centering
  \includegraphics[width=0.5\textwidth]{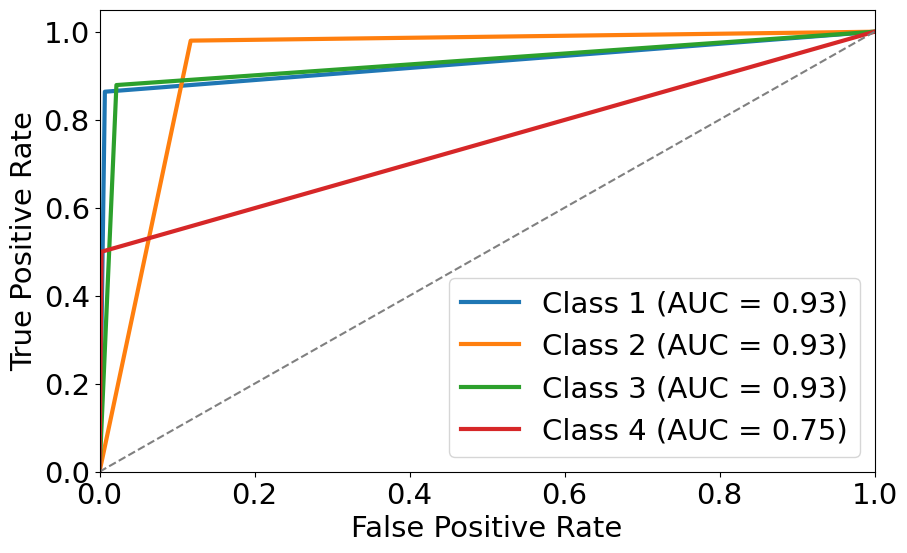}
  \caption{Multi-Class ROC Curve for the distress level prediction.}
  \label{fig:ROC}
\end{figure}

Figure \ref{fig:ROC} shows the AUC achieved by STGAN  in classifying different classes of road distress. Particularly,  it shows that our model can have quite good AUC values for predicting the data in classes 1, 2, and 3, while performs relatively worse for the data in class 4. Importantly, this class maintains an extremely low false positive rate, indicating minimal risk of false alarms in practical applications and preserving valuable human resources.
% Class 1 excels with an impressive AUC of 0.98, indicating exceptional effectiveness in distinguishing between positive and negative cases, boasting a high true positive rate and a low false positive rate. Class 2 also shows strong performance with an AUC of 0.95. Although it experiences occasional misclassifications with other classes, such occurrences are relatively rare. Class 3 attains a decent AUC of 0.87, reflecting good performance but with more misclassifications compared to Class 1 and Class 2. This class also has a higher likelihood of false negatives and false positives.

% Lastly, Class 4 achieves an AUC of 0.74, which is noticeably lower than the other classes but still better than random guessing. 

\begin{figure*}[ht]
    \centering
    \begin{subfigure}{0.49\textwidth} % 设置第一张图片占总宽度的45%
        \centering
        \includegraphics[width=\textwidth]{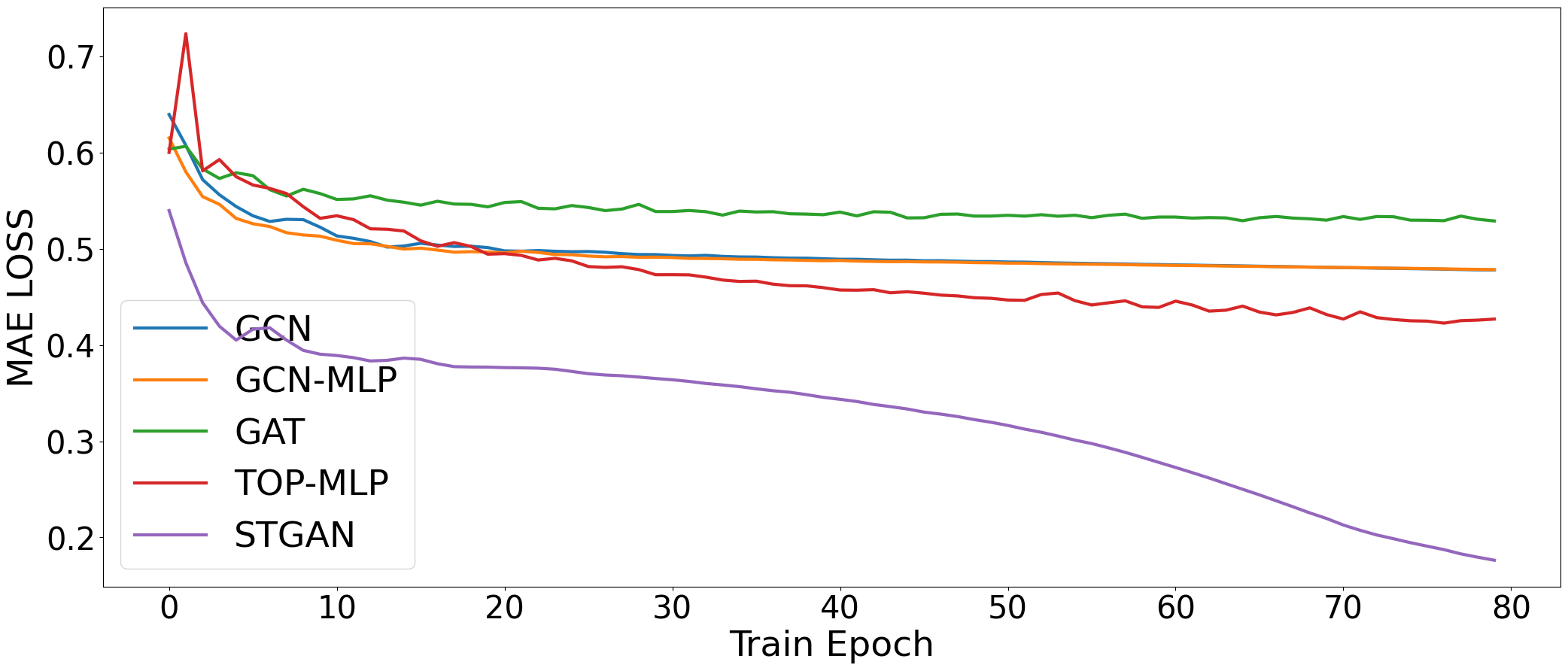}
        \caption{0-2000 Data}
    \end{subfigure}
    \begin{subfigure}{0.49\textwidth} % 设置第二张图片占总宽度的45%
        \centering
        \includegraphics[width=\textwidth]{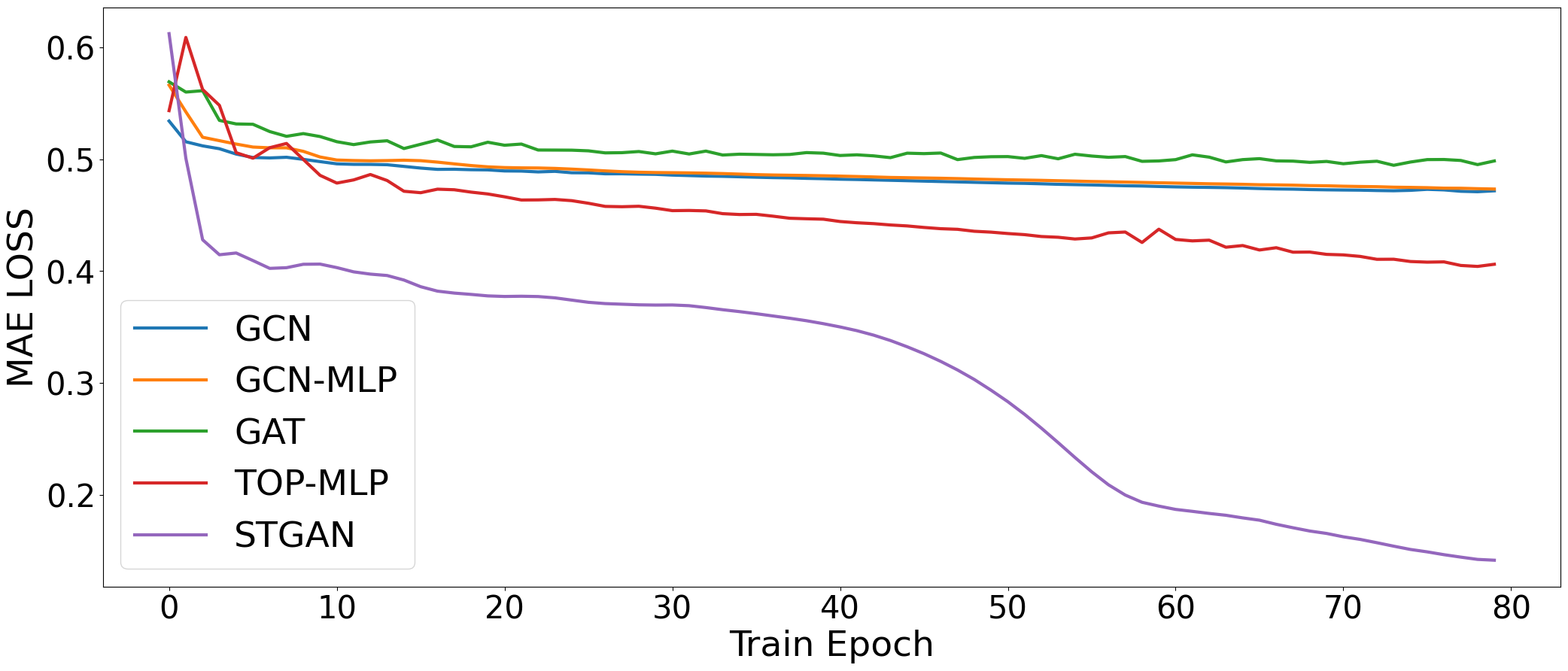}
        \caption{5000-7000 Data}
    \end{subfigure}
    \begin{subfigure}{0.49\textwidth} % 设置第三张图片占总宽度的45%
        \centering
        \includegraphics[width=\textwidth]{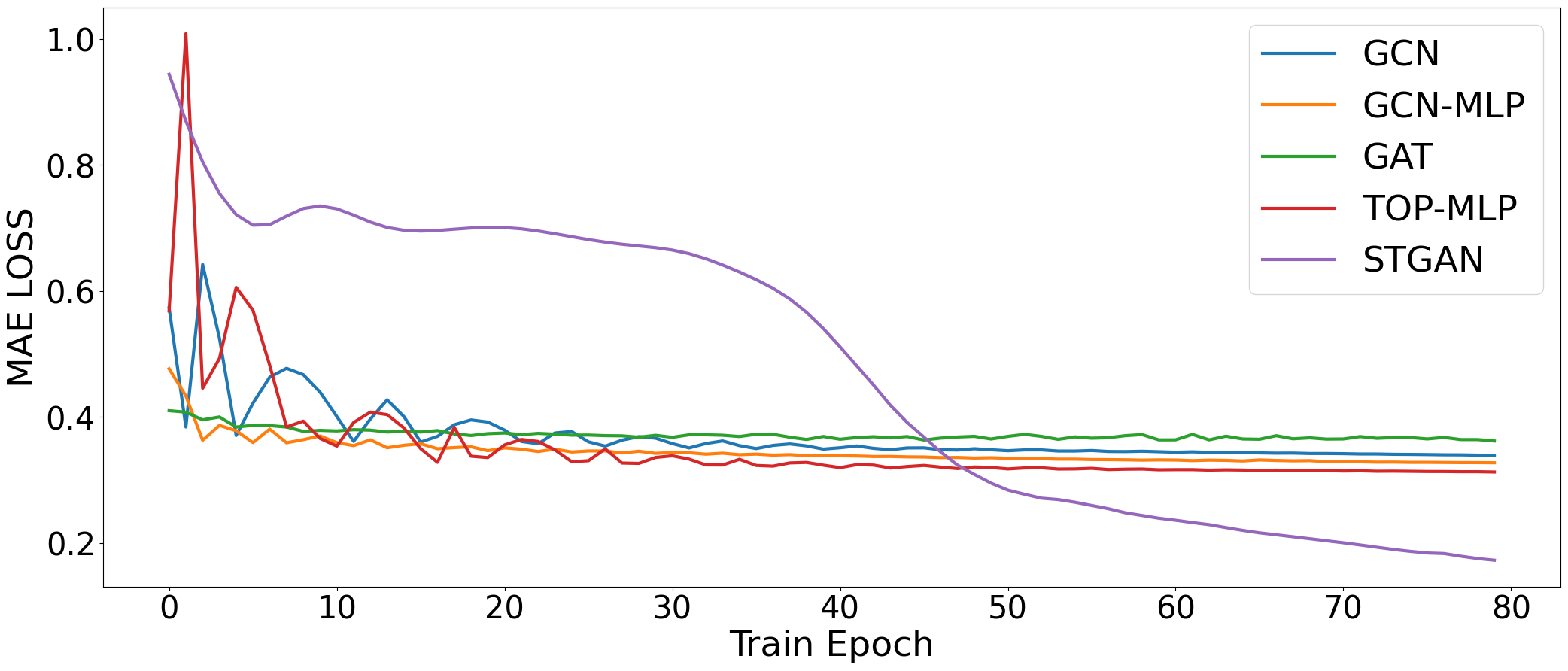}
        \caption{10000-12000 Data}
    \end{subfigure}
    \begin{subfigure}{0.49\textwidth} % 设置第四张图片占总宽度的45%
        \centering
        \includegraphics[width=\textwidth]{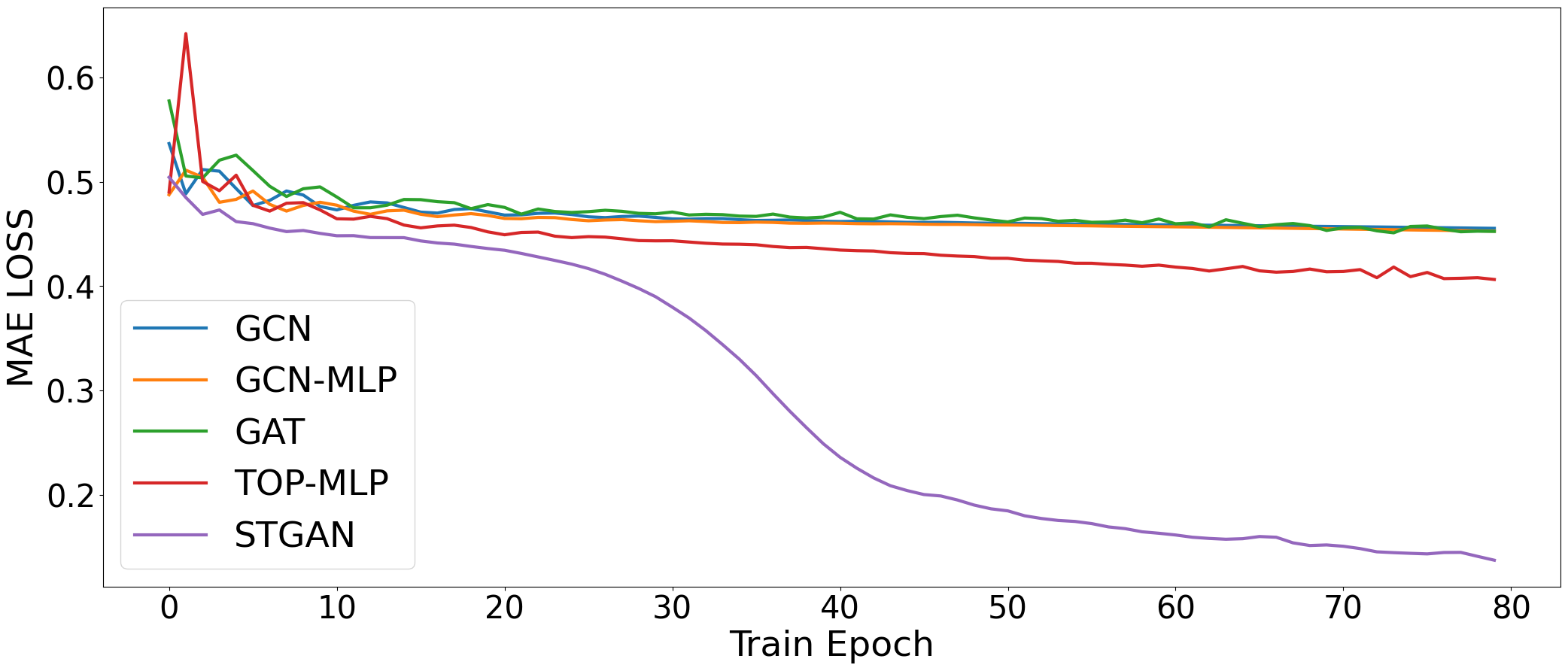}
        \caption{15200-17200 Data}
    \end{subfigure}
    
    \caption{Training loss comparison between STGAN and baseline methods at different periods. It can be seen that the developed STGAN model can uniformly achieve smaller training loss with sufficient training epochs.}
    \label{fig:STGAN_LOSS}
\end{figure*}
\subsection{Performance Comparisons with Baseline Models}

In this part, we conducted comparative experiments between STGAN and four baseline models: TOP-MLP, GCN, GCN-MLP, and GAT, by comparing the corresponding MAE, MSE, and RMSE on the test dataset. In particular, these four baseline models as well as STGAN are trained on the same training dataset with separately tuned hyperparameters.

Figure \ref{fig:STGAN_LOSS} displays the training loss curve for various models across four distinct data segments. Notably, STGAN's loss value is the lowest among all models, while the MAE values for the other four models are considerably higher, emphasizing STGAN's superior performance on the dataset. During STGAN's training process, there is a rapid decline in loss within the first 10 epochs, followed by a brief plateau phase, and another sharp decrease around the 60th epoch. As a result, we recommend training STGAN for no fewer than 60 epochs. 

In comparison, the training results reveal that GAT experiences the highest loss value after convergence, whereas TOP-MLP exhibits the lowest loss value post-convergence. Both GCN and GCN-MLP demonstrate similar performance levels. This emphasizes STGAN's enhanced performance compared to the baseline models in the training period.

\begin{figure*}[!tb]
    \centering
    \begin{subfigure}{0.49\textwidth} % 设置第一张图片占总宽度的45%
        \centering
        \includegraphics[width=\textwidth]{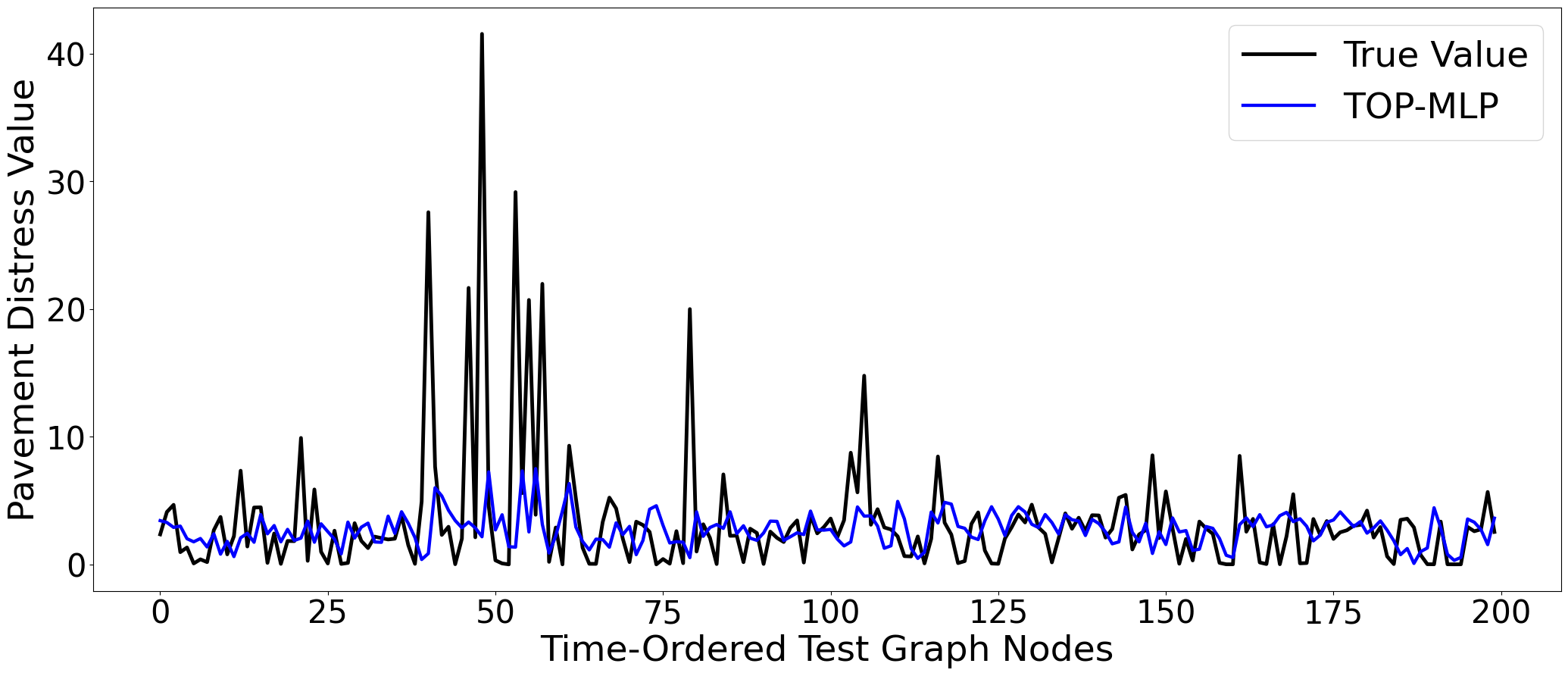}
        \caption{TOP-MLP Prediction Vs True Value}
    \end{subfigure}
    \begin{subfigure}{0.49\textwidth} % 设置第二张图片占总宽度的45%
        \centering
        \includegraphics[width=\textwidth]{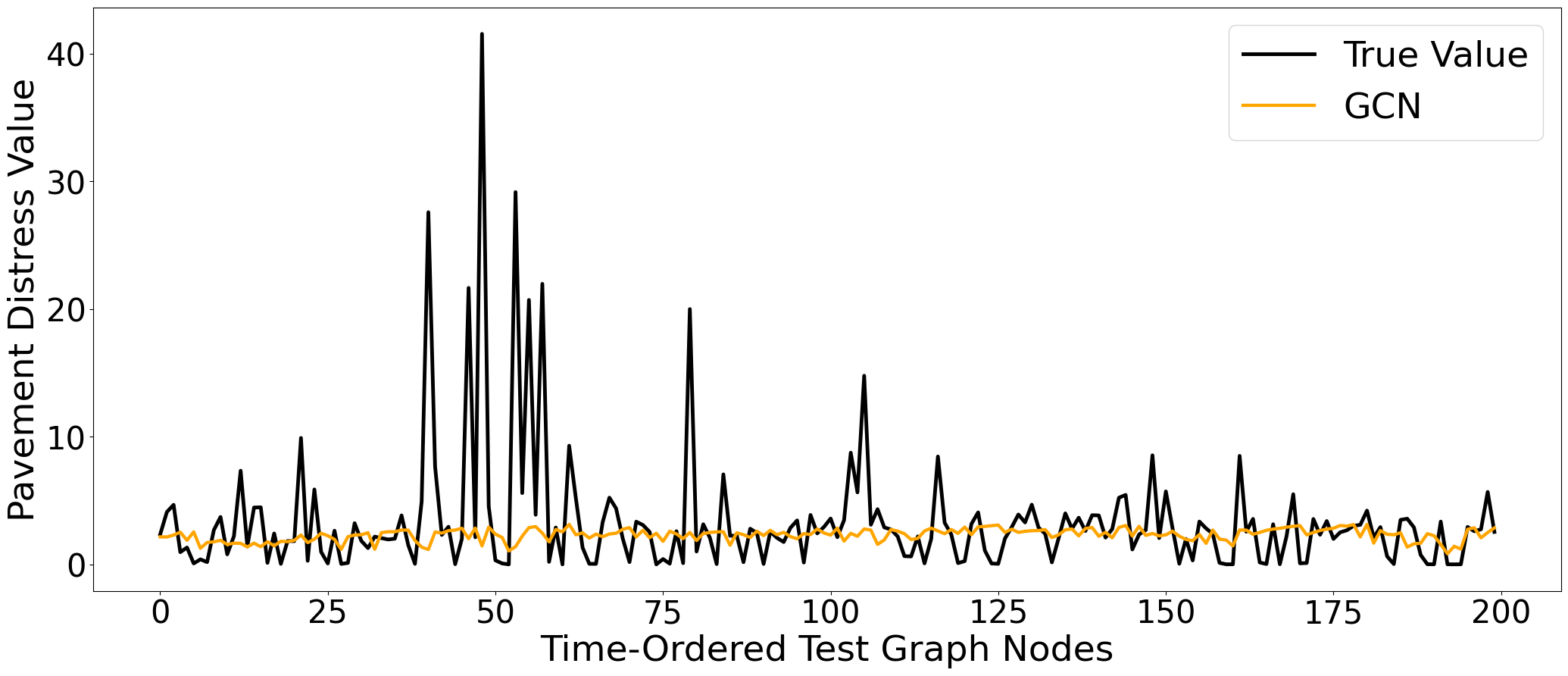}
        \caption{GCN Prediction Vs True Value}
    \end{subfigure}
    \begin{subfigure}{0.49\textwidth} % 设置第三张图片占总宽度的45%
        \centering
        \includegraphics[width=\textwidth]{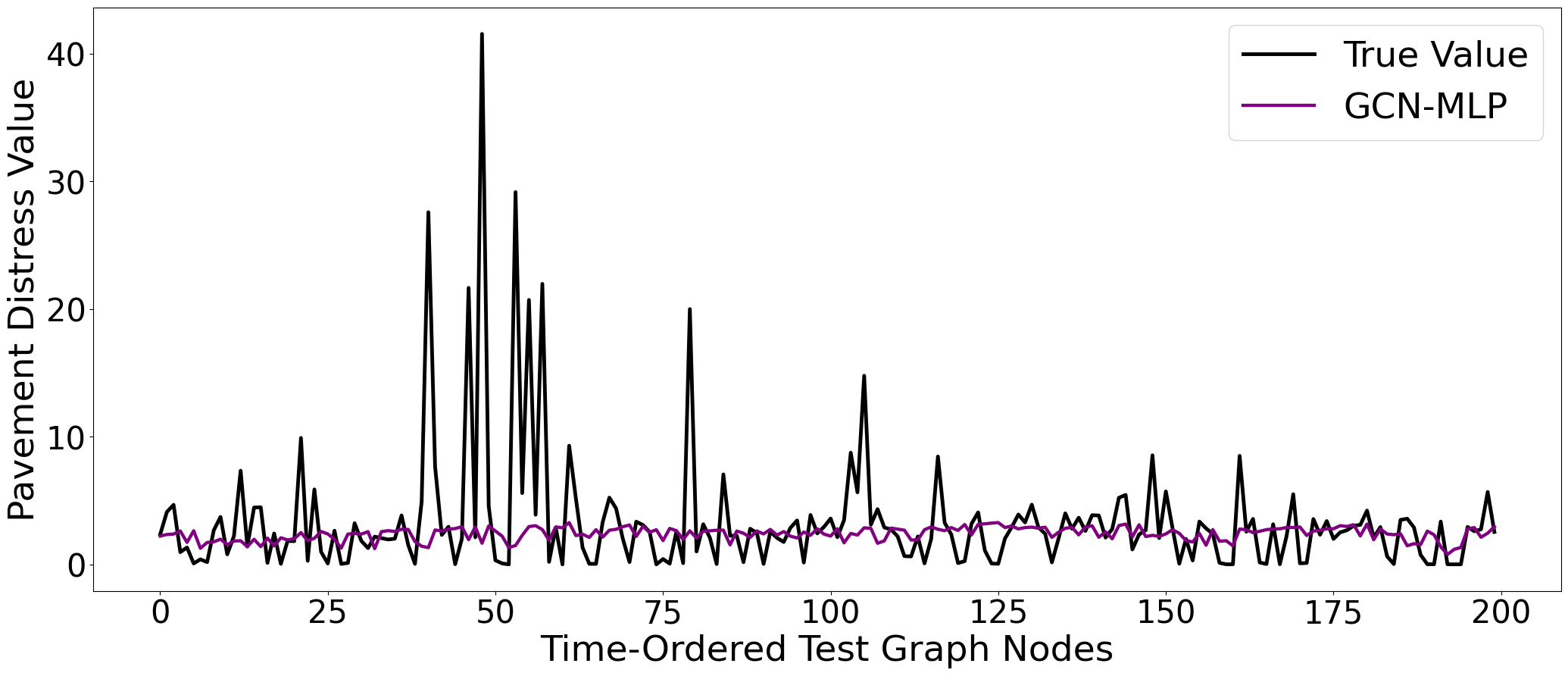}
        \caption{GCN-MLP Prediction Vs True Value}
    \end{subfigure}
    \begin{subfigure}{0.49\textwidth} % 设置第四张图片占总宽度的45%
        \centering
        \includegraphics[width=\textwidth]{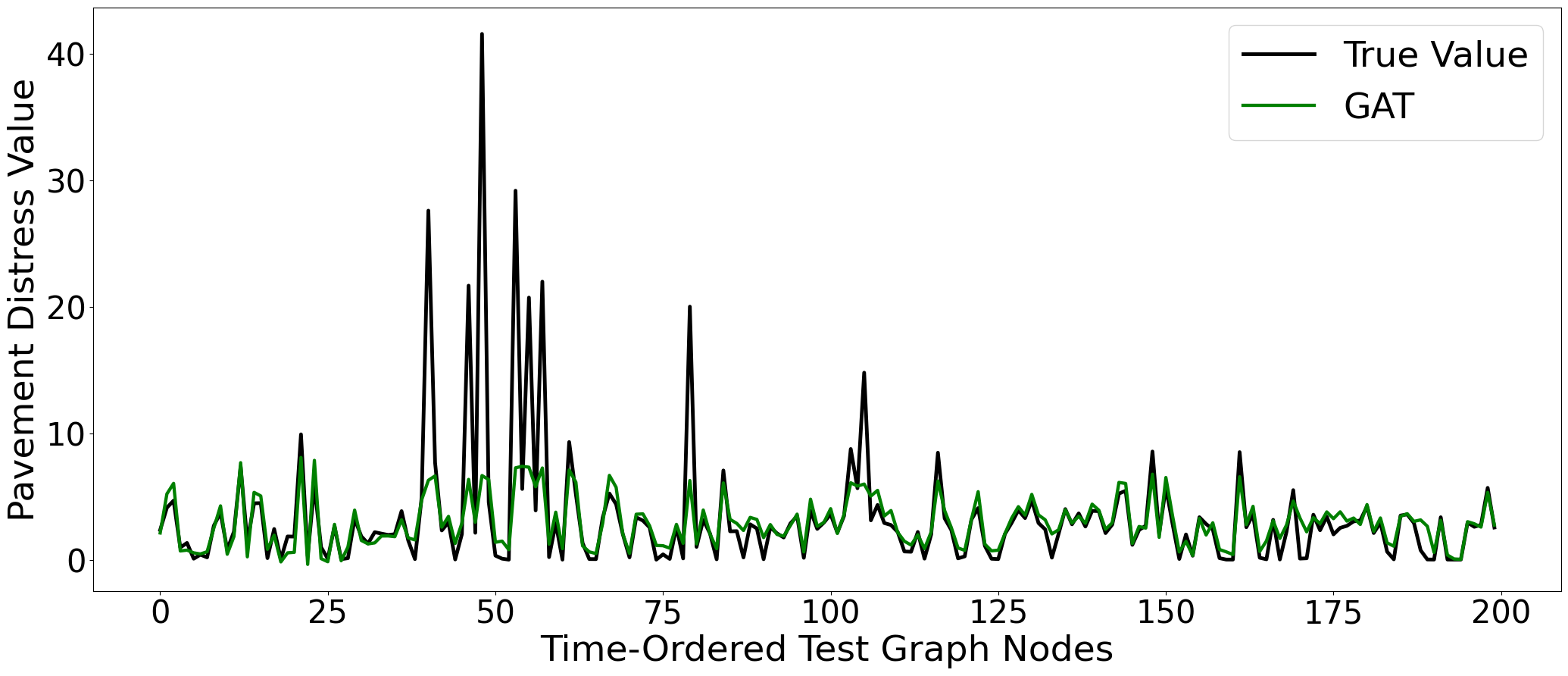}
        \caption{GAT Prediction Vs True Value}
    \end{subfigure}
    \caption{Performance of the baseline methods in the period $0-2000$. When comparing with the performance of STGAN shown in Figure \ref{fig:STGAN_pre_true}, these baseline methods clearly perform much worse. }
    \label{fig:models_pre_true}
\end{figure*}

The comparison between predicted and actual values for five different models on the 0-2000 data segment during the test period is illustrated in Figure \ref{subfig:STGAN_pre_true_a} and Figure \ref{fig:models_pre_true}. In summary, TOP-MLP can predict the general trend of changes, but cannot provide high precision. Its performance on the test set is noticeably worse than that on the training set, indicating a significant overfitting issue. Regarding GCN and GCN-MLP, both models can only predict partial data trends and struggle to predict extremely large or small values, resulting in predicted values that oscillate within a specific range. This may be caused by the smoothing effect of graph convolution operations using a fixed graph. Moreover, GAT performs much better than other baselines, it not only predicts the overall data trend but also captures subtle variations. However, its accuracy remains considerably lower compared to STGAN.

\begin{table*}[!t]
  \centering
  \caption{Test error comparison between STGAN and baseline models. }
  \label{tab:test_performance}
  \resizebox{\linewidth}{!}{
  \begin{tabular}{l|c|c|c|c|c|c|c|c|c|c|c|c}
    \hline
     \multirow{2}{*}{Data Segment} & \multicolumn{3}{c|}{0-2000} & \multicolumn{3}{c|}{5000-7000} & \multicolumn{3}{c|}{10000-12000} & \multicolumn{3}{c}{15200-17200} \\
     % \hline
    & MAE & MSE & RMSE & MAE & MSE & RMSE & MAE & MSE & RMSE & MAE & MSE & RMSE \\
    \hline
    TOP-MLP & 0.7129 & 2.0563 & 1.4340 & 0.4424 & 0.3797 & 0.6162 & 0.4211 & 0.3250 & 0.5701 & 0.4251 & 0.3048 & 0.5521 \\
    % \hline
    GCN & 0.6631 & 1.9274 & 1.3883 & 0.4198 & 0.3370 & 0.5805 & 0.4808 & 0.4089 & 0.6395 & 0.4494 & 0.3402 & 0.5833 \\
    % \hline
    GCN-MLP & 0.6739 & 1.9719 & 1.4042 & 0.4138 & 0.3334 & 0.5774 & 0.4710 & 0.3882 & 0.6231 & 0.4353 & 0.3250 & 0.5701\\
    % \hline
    GAT & 0.4609 & 1.4053 & 1.1855 & 0.3316 & 0.2434 & 0.4934 & 0.2417 & 0.1052 & 0.3243 & 0.1837 & 0.0852 & 0.2919\\
    % \hline
    STGAN & \textbf{0.1759} & \textbf{0.3673} & \textbf{0.6060} & \textbf{0.0918} & \textbf{0.0363} & \textbf{0.1905} & \textbf{0.1531} & \textbf{0.0566} & \textbf{0.2379} & \textbf{0.0719} & \textbf{0.0159} & \textbf{0.1259} \\
    \hline 
  \end{tabular}
  }
\end{table*}

Table \ref{tab:train_test_time} compares the training time and the prediction time for each model. TOP-MLP is the fastest due to its simple neural network structure, while STGAN takes about 30\% more time than GNN and GAT for training due to the need to extract spatiotemporal coefficients and time differences. However, the prediction time for STGAN is similar to that for GNN and GAT.

\begin{table}[!t]
  \centering
  \caption{Training and prediction time comparison between STGAN and baseline models.}
  \label{tab:train_test_time}
 
    \begin{tabular}{c|c|c|c|c}
    \hline
    \textbf{Model} & \textbf{TOP-MLP} & \textbf{GCN} & \textbf{GAT} & \textbf{STGAN} \\ 
    \hline
    \textbf{Train Time (s)} & 1.31 & 3.32 & 2.84 & 4.18 \\ 
    \textbf{Prediction Time (s)} & 0.01 & 0.15 & 0.25 & 0.22 \\ \hline
    \end{tabular}
      
\end{table}

By analyzing the performance of various models on the test data set shown in Table \ref{tab:test_performance}, we can draw the following four conclusions:
\begin{itemize}[leftmargin=*]
  \item In general, models that involve GNN outperform MLP-only models, highlighting the importance of modeling the data as graph structures.
  \item Networks with attention mechanisms, such as GAT and STGAN, outperform other models. This suggests that pure graph convolutional methods are insufficient to capture the differences between adjacent nodes, whereas attention mechanisms can selectively extract features from neighboring nodes, providing more informative and reasonable correlation between nodes and 
  leading to more accurate predictions.
  \item STGAN performs better than GAT, indicating that STGAN is more sensitive to the spatial-temporal correlations between nodes and captures them more accurately.
\end{itemize}

\subsection{Ablation Experiment}
In this subsection, we will perform several experiments to further evaluate the importance of three key modules in the STGAN model: the TOP connection, spatial-temporal attention mechanism, and time difference feature. 
The definitions of these models will be added later.

% \begin{table}[htb]
%   \centering
%   \caption{Ablation Experiment Results}
%   \label{tab:ablation}
%   \resizebox{\linewidth}{!}{
%   \begin{tabular}{c|c|c|c|c|c|c|c|c|c|c|c|c}
%     \hline
%     \multirow{2}{*}{Data Segment} & \multicolumn{3}{c|}{0-2000} & \multicolumn{3}{c|}{5000-7000} & \multicolumn{3}{c|}{10000-12000} & \multicolumn{3}{c}{15200-17200} \\
%      % \hline
%     & MAE & MSE & RMSE & MAE & MSE & RMSE & MAE & MSE & RMSE & MAE & MSE & RMSE \\
%     \hline
%     STGAN & \textbf{0.1759} & \textbf{0.3673} & \textbf{0.6060} & \textbf{0.0918} & \textbf{0.0363} & \textbf{0.1905} & \textbf{0.1531} & \textbf{0.0566} & \textbf{0.2379} & \textbf{0.0719} & \textbf{0.0159} & \textbf{0.1259} \\
    
%     % \hline 
%     STGAN w/o TOP & 0.6717 & 1.8686 & 1.3670 & 0.4293 & 0.3474 & 0.5894 & 0.4851 & 0.4337 & 0.6585 & 0.4192 & 0.3192 & 0.5650 \\
    
%     % \hline
%     GATST & 0.7130 & 2.0564 & 1.4340 & 0.4424 & 0.3798 & 0.6163 & 0.4211 & 0.3250 & 0.5701 & 0.4251 & 0.3048 & 0.5522 \\
    
%     % \hline 
%     STGAN w/o TD feature & 0.6631 & 1.9275 & 1.3883 & 0.4198 & 0.3370 & 0.5805 & 0.4808 & 0.4089 & 0.6395 & 0.4494 & 0.3402 & 0.5833 \\
%     \hline 
%   \end{tabular}
%   }
% \end{table}

\subsubsection{TOP Connection}
The TOP connection serves to complete the information. For discussion on the significance of TOP connection mechanism, we conducted experiments named \textbf{STGAN Without TOP} (STGAN w/o TOP), which does not utilize the TOP connection mechanism and only employs the natural connection to construct the graph. The experimental results indicate that the performance of STGAN-WT is significantly inferior to STGAN. This discrepancy is attributed to the role played by the TOP mechanism in information completion. Some nodes with few or even no natural connections can benefit from TOP connection channels of information, thereby enhancing their predictive capabilities. And for nodes with many natural connections, the TOP mechanism also serves as a supplementary factor. More importantly, under the influence of attention mechanisms, there is no concern that TOP connections will dominate, as the model still prioritizes information from natural connections. Therefore, the inclusion of the TOP connection mechanism results in a significant improvement in the model's capabilities.

% \hspace*{\fill}

% As for selecting the number of TOP connections, an excessive quantity leads to decreased performance, due to the interference caused by an abundance of irrelevant information, which hampers the model's predictive abilities. Conversely, if the number is too low, the performance also deteriorates because insufficient information completion can occur.

\subsubsection{Spatial-temporal Attention Mechanism}

The spatiotemporal attention mechanism significantly enhances the performance of the STGAN model. Its significance has been demonstrated through comparative experiments with GAT, which utilizes the regular attention mechanism and all features for it. The phenomenon that STGAN performs better in comparative experiments is because we intentionally select the most important spatiotemporal information as the input for calculating the attention coefficient matrix, allowing the model to effectively filter out interference from other information. 

In that subsection, in order to discuss the optimal implementation of the spatiotemporal attention mechanism, we introduced another spatial-temporal attention mechanism, \textbf{STGAN with the explcit attention mechanism} (STGAN w/ EAM), that leverages the spatial and  temporal discrepancies between two nodes explicitly. Assuming the spatial and temporal difference between two points is represented as $\Delta l_{ij}$ and $\Delta t_{ij}$ respectively, the attention mechanism adopts a GAT network based on all features to get an initial coefficient $a_{ij}$, and then aggregating the spatial and temporal differences through a Softmax layer to obtain the final spatial-temporal attention coefficients.  
\begin{align*}
w_{ij} = \mathrm{softmax}\big(a_{ij} \cdot \exp(-\gamma \Delta l_{ij}) \cdot \exp(-\gamma \Delta t_{ij})\big),
\end{align*}
where $\gamma>0$ is a tuning parameter and we consider exponential decaying function to spatial and temporal difference. The STGAN model with such a different attention mechanism performs significantly worse than the original STGAN. This shows that directing using spatiotemporal features to form attention coefficients, which is utilized in STGAN, is superior to aggregating all features with explicit spatial and temporal differences to obtain attention coefficients. Therefore, explicitly emphasizing the importance of spatiotemporal information in the feature space is more effective in enabling the model to focus on spatiotemporal relationships.

\subsubsection{Time Difference Feature}
The time difference feature is primarily used as an input for calculating the attention coefficients matrix, whose role is to explicitly highlight the significant impact of time difference on spatiotemporal correlations, rather than simply inputting nodes' time information features. We designed an experiment called \textbf{STGAN without time difference feature} (STGAN w/o TD feature), where we removed the input of time difference information during the calculation of attention coefficient matrix. The experimental results indicate that STGAN outperforms this new model by a significant margin. The use of time difference leads to a substantial performance improvement, because without utilizing the time difference, the model would need to find it between nodes' standardized time features, which is less direct than simply providing it.

Then the comparison resutls are shown in Table \ref{tab:ablation}. It clearly demonstrates the importance of TOP mechanism, our attention calculation methods, and time difference features.

\begin{table}[!tb]
	\centering
	\caption{Ablation studies on the TOP mechanism, attention calculations, and time difference features in STGAN. Errors are measured as MAE. Here periods 1, 2,3,4 stand for the data segments 0-2000, 5000-7000, 10000-12000, and 15200-17200 respectively.}
	\label{tab:ablation}
	\begin{tabular}{c|c|c|c|c}
		\hline
		Data Segment & Period 1 & Period 2 & Period 3 & Period 4 \\
		% \hline
		\hline
		
		% \hline 
		STGAN w/o TOP & 0.6717 & 0.4293 & 0.4851 & 0.4192  \\
		
		% \hline
		STGAN w/  EAM & 0.7130 & 0.4424 & 0.4211 & 0.4251 \\
		
		% \hline 
		STGAN w/o TD feature & 0.6631 & 0.4198 & 0.4808 & 0.4494\\
		\hline 
	\end{tabular}
\end{table}

\subsubsection{Environmental Data}
\begin{figure}[!tb]
\centering
\hspace{-0.5cm}
\includegraphics[width=0.5\textwidth]{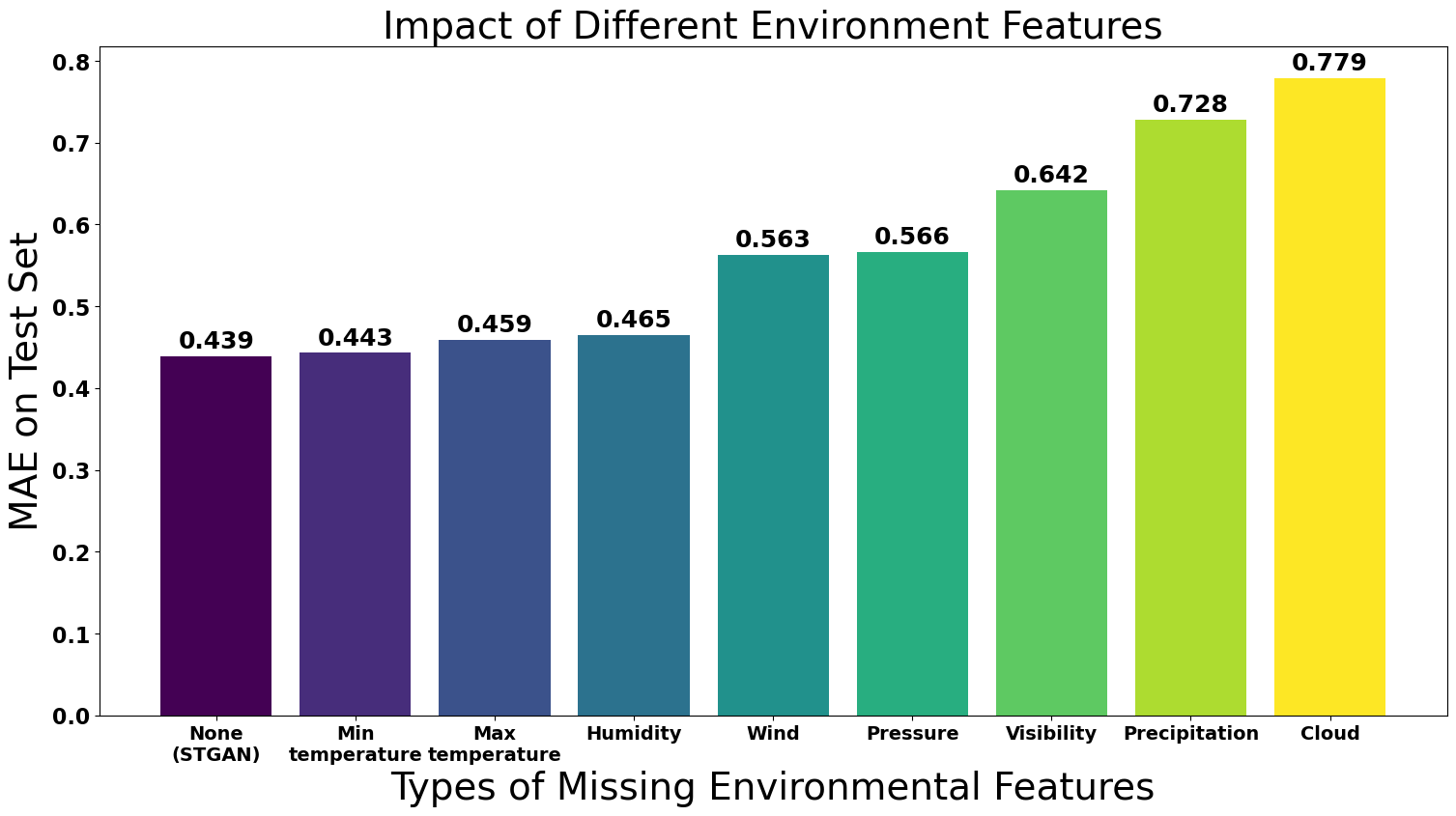}
    \centering
    \caption{Impact of different environment factors.}
    \label{fig:env_difff_bar}
\end{figure}

We conducted an ablation experiment on the environmental data across eight dimensions by masking one specific dimension at a time. Specifically, we removed each dimension from the input features of our model to evaluate the performance loss compared to the proposed STGAN model. This approach helps assess the importance of each environmental factor in relation to the prediction performance. The results are summarized in Figure \ref{fig:env_difff_bar}. It is evident that removing any environmental feature leads to performance loss, with different factors demonstrating varying levels of importance. Notably, visibility, precipitation, and cloud cover are the top three most important factors for prediction performance, while minimum/maximum temperatures and humidity have a lesser impact on the model's performance.

\subsection{Generalization Analysis}

\begin{figure}[!tb]
\centering
\includegraphics[width=0.5\textwidth]{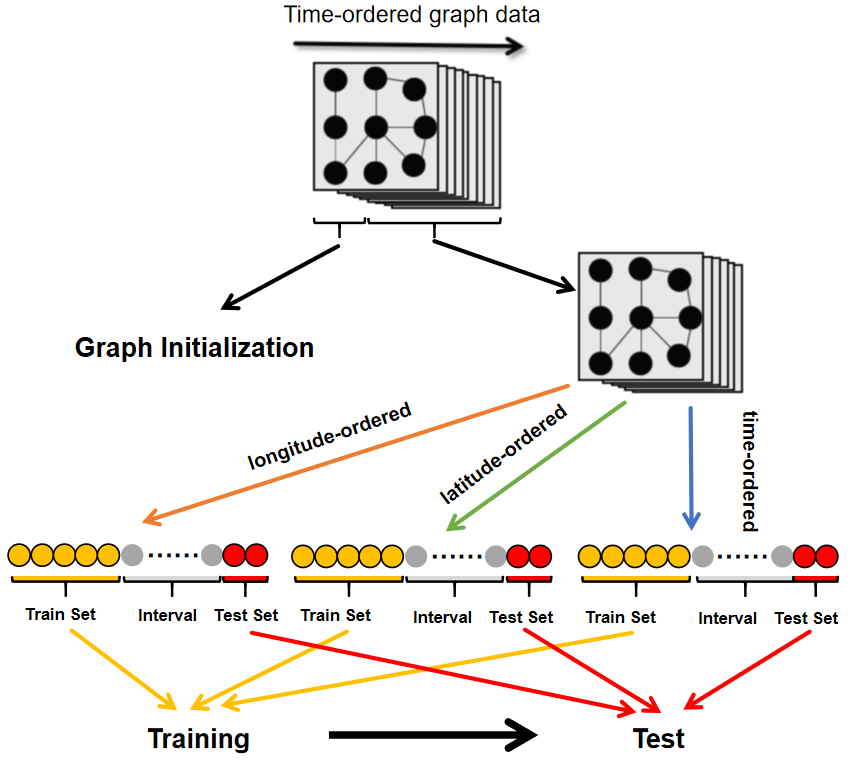}
    \centering
    \caption{The diagram of a method for the data splitting in both time and spatial domains.}
    \label{fig:howToSplit}
\end{figure}

We further investigate the generalization ability of the proposed model, specifically focusing on its ability to generalize to different regions and time periods. Due to data collection limitations, we cannot obtain data from regions other than Shanghai. Therefore, we divide Shanghai into two regions based on two dimensions: longitude and latitude. For longitude-based separation, we first sort the locations (i.e., nodes) by their longitude and then perform the separation accordingly. A similar approach is applied for latitude and timestamps. To further assess the region-generalization ability, we consider different separation levels by removing a certain number of data points between the two regions, denoted as the interval number. For instance, let $x_1, \dots, x_n$ represent the sorted data points. We remove the data from $x_{k+1}$ to $x_{k+s}$ to ensure that the training set $x_1, \dots, x_k$ and the test set $x_{k+s+1}, \dots, x_n$ have sufficient separation. The splitting method is summarized in the diagram shown in Figure \ref{fig:howToSplit}.

% \textcolor{red}{The experimental results are shown in Figure \ref{fig:}
% The discussion of the input spatiotemporal data will be conducted in two dimensions. The first dimension involves different sorting orders, specifically sorting by longitude, latitude, and time, to explore the impact of sorting order on model performance. When sorted by longitude, this essentially divides the regions by longitude, meaning that the training and testing sets do not belong to the same region. The same logic applies when sorting by latitude. The second dimension examines the effect of different intervals after sorting, where the number of data points between the training and testing sets varies. This allows for an analysis of how the spatial and temporal spacing influences model performance. An illustrative diagram, shown in Figure \ref{fig:howToSplit}, demonstrates that after sorting all the data points by time, a small subset is used for the initialization of the graph, while the majority of points are sorted by longitude, latitude, or time. A specific number of data points are spaced between the training and testing sets.}

\begin{figure}[!tb]
\centering
\includegraphics[width=0.5\textwidth]{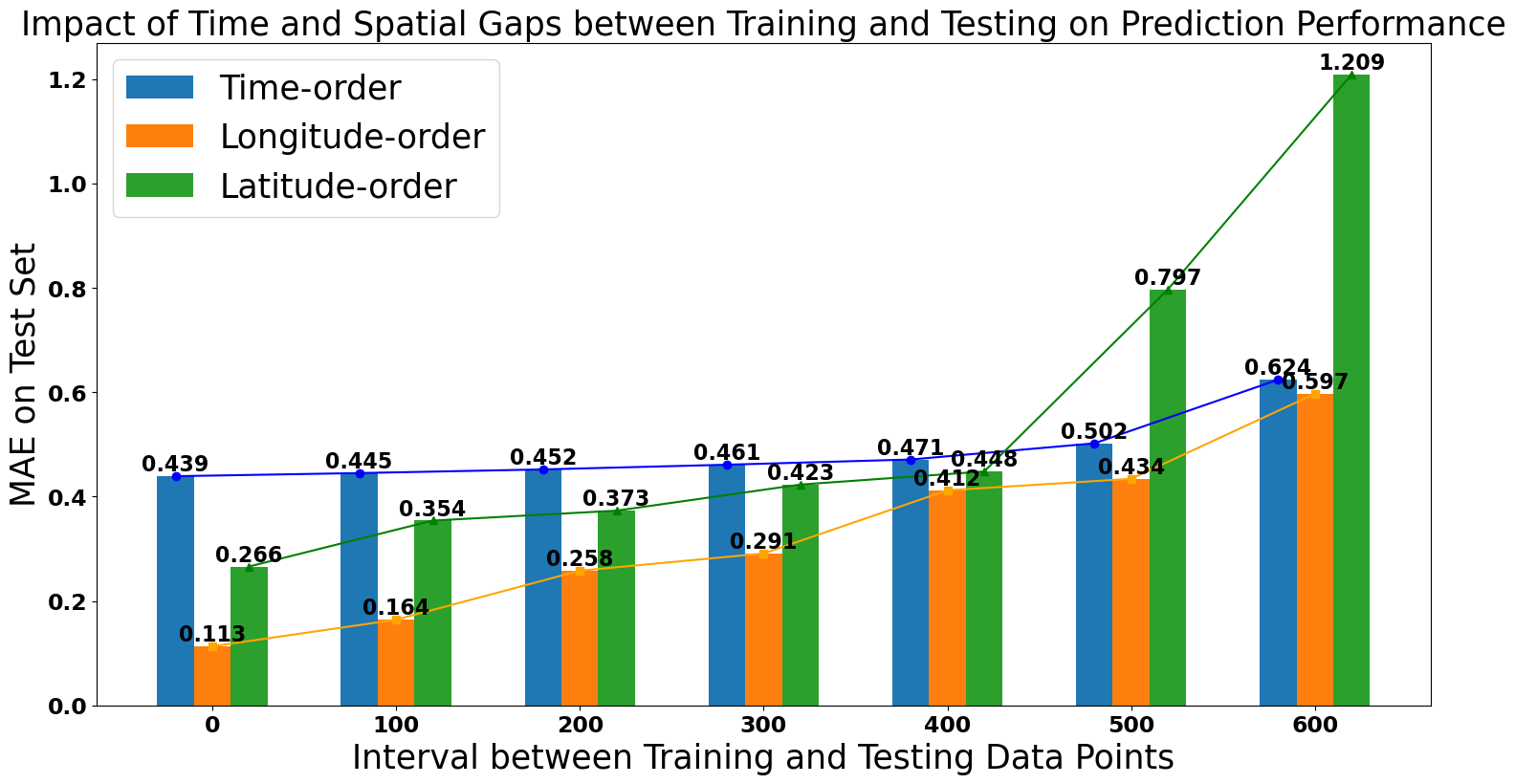}
    \centering
    \caption{Impact of Time and Spatial Gaps Between Test and Training Data on Prediction Performance}
    \label{fig:time_space_gap}
\end{figure}

The results are depicted in Figure \ref{fig:time_space_gap}, where different colors represent various data sorting sequences. Blue indicates time order, orange represents dimension order, and green signifies longitude order. The vertical axis shows the model's MAE on the test dataset, while the horizontal axis displays the interval between data points in the test and training datasets for each sorting scheme.
It is evident that when the separation interval is small, the region-level generalization is reasonably good, whereas the time-level generalization can be poorer. Conversely, with a larger separation interval, there is a significant increase in prediction error for region-level generalization, while the performance for time-level generalization remains relatively stable. This suggests that (1) it is crucial to ensure that the training data is up to date, and (2) the model can adapt to different regions, provided they are not significantly distant from the regions of the training data points.

% \textcolor{red}{Furthermore, regardless of the interval point count, the model trained on the dataset sorted by dimension produces the best results. Longitude sorting follows, with time sorting yielding the poorest model performance. While this is largely due to the data being collected along a specific route, where both dimension and longitude sorting inherently carry some temporal information, this phenomenon also highlights the importance of spatial factors to some extent.}

\subsection{Hyperparameter Tuning and Comparative Analysis}

% \begin{table}[htb]
%   \centering
%   \caption{Hyperparameter Tuning Experiment Results}
%   \label{tab:hyperpara}
%   \resizebox{\linewidth}{!}{
%   \begin{tabular}{l|c|c|c|c|c|c|c|c|c|c|c|c}
%     \hline
%     \multirow{2}{*}{Data Segment} & \multicolumn{3}{c|}{0-2000} & \multicolumn{3}{c|}{5000-7000} & \multicolumn{3}{c|}{10000-12000} & \multicolumn{3}{c}{15200-17200} \\
%      % \hline
%     & MAE & MSE & RMSE & MAE & MSE & RMSE & MAE & MSE & RMSE & MAE & MSE & RMSE \\
%     \hline
%     STGAN & \textbf{0.1759} & \textbf{0.3673} & \textbf{0.6060} & \textbf{0.0918} & \textbf{0.0363} & \textbf{0.1905} & \textbf{0.1531} & \textbf{0.0566} & \textbf{0.2379} & \textbf{0.0719} & \textbf{0.0159} & \textbf{0.1259} \\
    
%     % \hline 
%     STGAN (H=1) & 0.3777 & 0.8074 & 0.8985 & 0.2869 & 0.2053 & 0.4531 & 0.9110 & 1.7126 & 1.3086 & 0.2986 & 0.1868 & 0.4321 \\
%     % \hline 
%     STGAN (H=10) & \textbf{0.1538} & \textbf{0.2870} & \textbf{0.5357} & \textbf{0.0663} & \textbf{0.0360} & \textbf{0.1896} & \textbf{0.1697} & \textbf{0.0462} & \textbf{0.2150} & \textbf{0.0646} & \textbf{0.0108} & \textbf{0.1038} \\ 
%     % \hline 
%     STGAN (L=2) & 0.2737 & 0.4026 & 0.6345 & 0.2909 & 0.1716 & 0.4143 & 0.5407 & 0.5311 & 0.7288 & 0.2838 & 0.1509 & 0.3884 \\ 
%     % \hline 
%     STGAN (L=3) & 0.3573 & 0.6174 & 0.7857 & 0.2386 & 0.1478 & 0.3844 & 0.8815 & 1.5394 & 1.2407 & 0.3826 & 0.2391 & 0.4890 \\ 
%     \hline 
%   \end{tabular}
%   }
% \end{table}

\begin{figure}[!t]
  \begin{minipage}{0.24\textwidth}
    \centering
    \includegraphics[width=\linewidth]{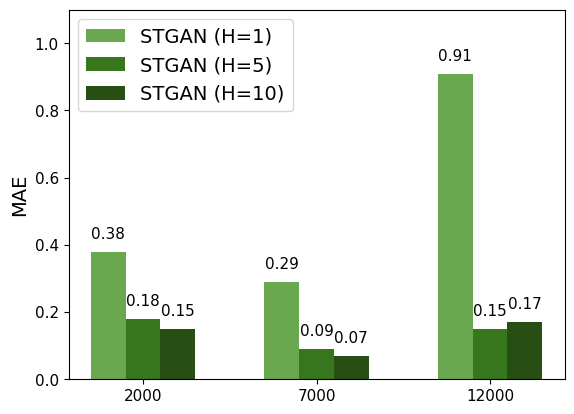}
    \caption{MAE Change with different \# head.}
    \label{fig:Head_Hyper}
  \end{minipage}%
  \begin{minipage}{0.24\textwidth}
    \centering
    \includegraphics[width=\linewidth]{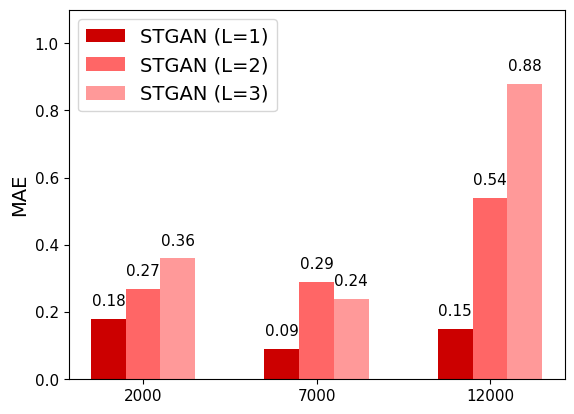}
    \caption{MAE Change with Different \# layer.}
    \label{fig:C_Hyper}
  \end{minipage}
\end{figure}

To discuss the impact of the number of attention heads and convolutional layers on the predictive performance of STGAN, we conducted two sets of experiments. 

\paragraph{The number of attention Heads} The first set involved altering only the number of attention mechanism heads, changing them from the original 5 heads to a single head and 10 heads, marked as $H=5$, $H=1$ and $H=10$ respectively. 

\paragraph{The number of convolutional layers} The second set involved modifying only the convolutional layers of the original model, going from a single convolutional layer to two and three layers of convolution, marked as $L=1$, $L=2$ and $L=3$ respectively. The multiple convolution operation was similar to GCN, except that we replaced the standard Laplacian matrix used in each convolution with an spatio-temporal attention coefficient matrix, and the attention coefficient matrix used in each convolution was the same. 

Figures \ref{fig:Head_Hyper} and \ref{fig:C_Hyper}  illustrate the ablation studies over the architecture hyperparameters of STGAN, from which we can observe:
\begin{itemize}
\item As the number of attention heads increases, the model's accuracy improves. However, when the number of heads become even larger (e.g., $>5$) the further improvements are marginal.

\item With an increase in the number of convolutional layers, the model's performance gradually deteriorates, consistent with the behavior of GCN models. This suggests that aggregating more points' information with multiple convolutional layers does not lead to performance improvement. Instead, it can confuse information and lead to decreased performance.
\end{itemize}

\section{Limitation and Discussion}
\begin{figure}[!th]
    \centering
    \begin{subfigure}{0.24\textwidth} % 设置第一张图片占总宽度的45%
        \centering
        \includegraphics[width=\textwidth]{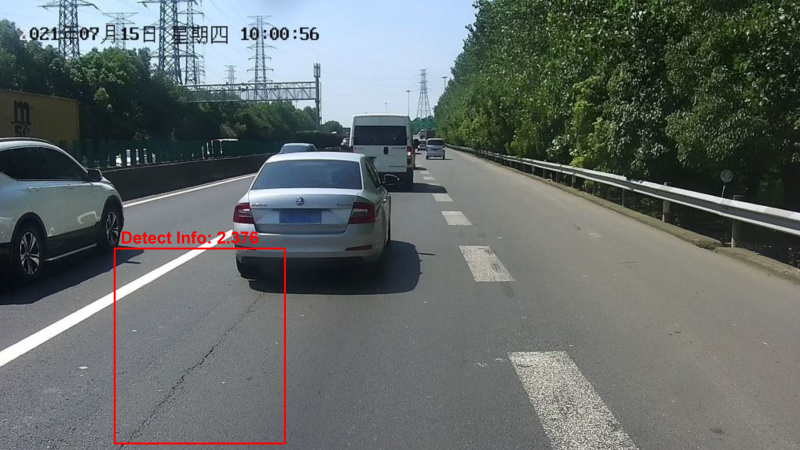}
        \caption{Same road crack with 2.376 DetectInfo value}
        \label{subfig:subfigure_a}
    \end{subfigure}
    \begin{subfigure}{0.24\textwidth} % 设置第二张图片占总宽度的45%
        \centering
        \includegraphics[width=\textwidth]{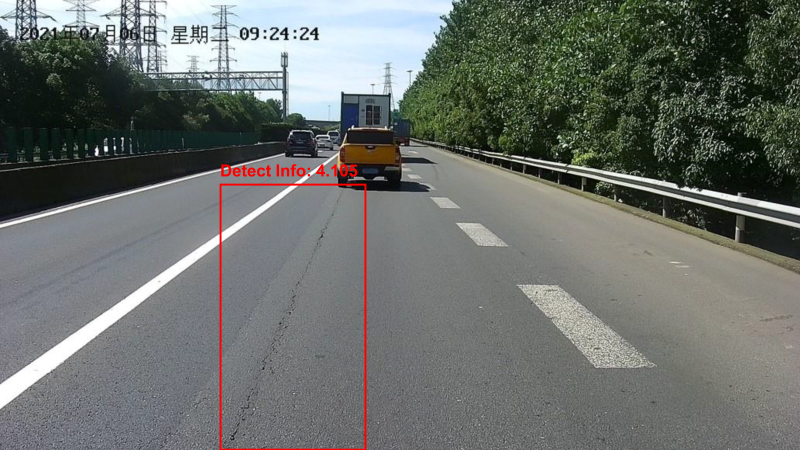}
        \caption{Same road crack with 4.105 DetectInfo value}
        \label{subfig:subfigure_b}
    \end{subfigure}
    \begin{subfigure}{0.24\textwidth} % 设置第三张图片占总宽度的45%
        \centering
        \includegraphics[width=\textwidth]{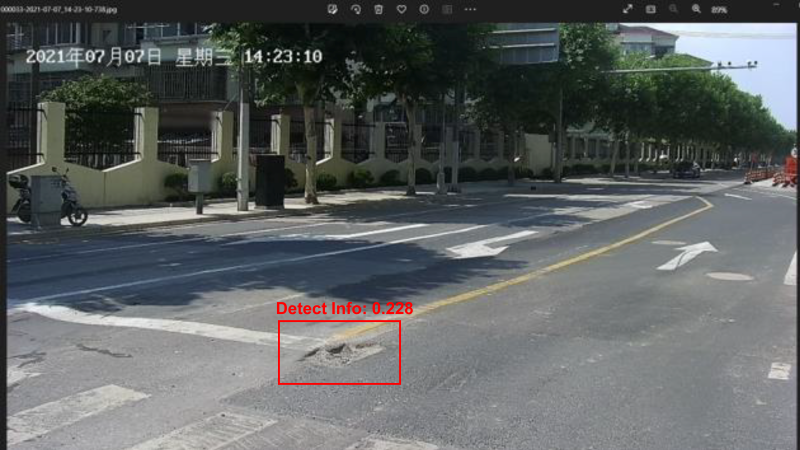}
        \caption{Small pothole with big DetectInfo value}
        \label{subfig:subfigure_c}
    \end{subfigure}
    \begin{subfigure}{0.24\textwidth} % 设置第四张图片占总宽度的45%
        \centering
        \includegraphics[width=\textwidth]{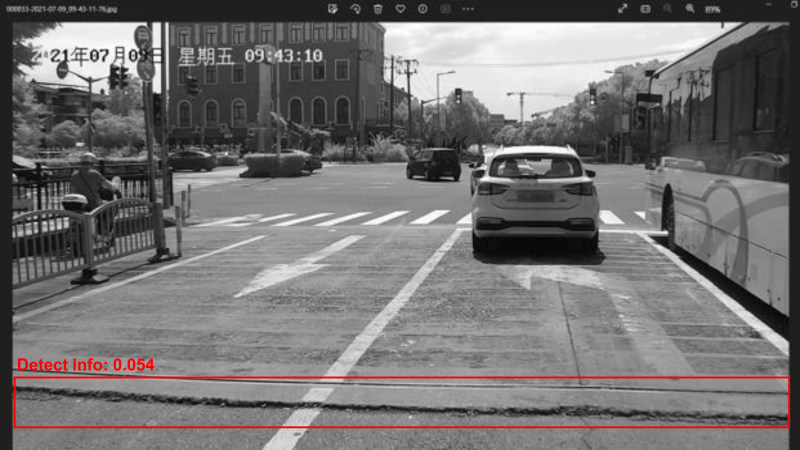}
        \caption{Big patch-crack with small DetectInfo value}
        \label{subfig:subfigure_d}
    \end{subfigure}
    \label{fig:Dataset_Issue}
    \caption{Contrack Dataset Issue and Limitation}
\end{figure}

% The Contrack dataset exhibits 3 main issues that constrain the performance of STGAN. 
\paragraph{Uneven distribution of high road fault value} The pavement distress image data is collected by data collection vehicles driving along predefined routes. There is an uneven distribution of high road fault value data within different time range, which results in our model lack of the capability to predict high road fault values and leads to significant variations in MAE when testing on different data segments extracted from time sequences.

\paragraph{Environmental factors affecting road distress recognition} As illustrated in Figures \ref{subfig:subfigure_a} and \ref{subfig:subfigure_b}, road cracks at the same location that appear visually similar can exhibit significant differences in detect info values (detect info value is another name of road fault value in dataset). These abnormal and unusual variations due to environmental problems are the primary reason for the model's low accuracy.

\paragraph{inconsistencies between visual appearance and detect info values} As depicted in Figures \ref{subfig:subfigure_c} and \ref{subfig:subfigure_d}, some potholes may appear severe but have low detect info values, while some patch-crack may seem unproblematic but exhibit high fault values. This discrepancy between detect info values and real-world conditions has the potential to significantly decrease the practicality of the model.

\section{Conclusion and Future Work}\label{sec:conclusion}

In this paper, we formulated pavement distress deterioration prediction as a spatiotemporal graph autoregression problem and proposed the \textit{spatiotemporal graph autoregression network} to address the challenges posed by uneven, asynchronous and sparse spatial-temporal data. Specifically, Our approach effectively constructs spatiotemporal graph by integrating the temporal domain into the spatial domain with natural and TOP connections. We employed two types of feature extractions to address the inconsistency between target and historical data, and further enhanced feature processing performance through an encoder-decoder architecture. Utilizing spatial-temporal attention mechanisms and graph convolution, STGAN aggregates information from neighboring nodes and captures spatiotemporal correlations effectively. When evaluated on the real-world Contract dataset, STGAN demonstrates significantly better predictions compared to baseline models, particularly in the early stages of deterioration. Experimental results demonstrate that STGAN outperforms baseline models, the ablation studies verify the efficacy and necessity of the developed modules. Future research will be considered from two aspects: data and models. First, a pavement distress dataset with stronger certainty and finer granularity will be constructed, which support us to perform multi-scale fusion \cite{zang2018long} for further performance improvements. Factors such as traffic load, pavement maintenance, and geographical features will be taken into account to optimize the performance of the prediction model. On the other hand, efforts will be made to explore a better architecture that can more effectively utilize the features and reduce the computational cost when the sample size of pavement distress data is relatively small.

\bibliographystyle{IEEEtran}
\bibliography{reference}

\end{document}